\documentclass{article} % For LaTeX2e
\usepackage{iclr2026_conference,times}

\usepackage{amsmath,amsfonts,bm}

\def\eqref#1{equation~\ref{#1}}
\def\1{\bm{1}}

\DeclareMathAlphabet{\mathsfit}{\encodingdefault}{\sfdefault}{m}{sl}
\SetMathAlphabet{\mathsfit}{bold}{\encodingdefault}{\sfdefault}{bx}{n}

\usepackage{hyperref}
\usepackage{graphicx}
\usepackage{url}
\usepackage{booktabs} % For professional tables
\usepackage{listings}
\usepackage{xcolor}
\usepackage{pifont}
\usepackage{wrapfig}
\usepackage{wasysym}
\usepackage{tcolorbox}
\usepackage{amsmath}
\usepackage{multirow}
\usepackage{amsfonts}
\usepackage{enumitem} % 放到导言区

\title{AXIS: Explainable Time Series Anomaly Detection with Large Language Models}
\newcommand{\methodname}{AXIS}

\author{
Tian Lan \quad Hao Duong Le \quad Jinbo Li \\
\texttt{\{lant23, lijb22\}@mails.tsinghua.edu.cn} \\
\texttt{Lehaoduong.Vn@gmail.com} \\
\And
Wenjun He \quad Meng Wang \\
\texttt{\{hewenjun8, wangmeng71\}@huawei.com} \\
\AND
Chenghao Liu\thanks{Corresponding author.} \quad Chen Zhang\footnotemark[1] \\
\texttt{twinsken@gmail.com} \\
\texttt{chenzhang01@tsinghua.edu.cn}
}

\iclrfinalcopy % Uncomment for camera-ready version, but NOT for submission.
\begin{document}
\maketitle

\begin{abstract}
Time-series anomaly detection (TSAD) increasingly demands explanations that articulate not only if an anomaly occurred, but also what pattern it exhibits and why it is anomalous. Leveraging the impressive explanatory capabilities of Large Language Models (LLMs), recent works have attempted to treat time series as text for explainable TSAD. However, this approach faces a fundamental challenge: LLMs operate on discrete tokens and struggle to directly process long, continuous signals. Consequently, naïve time-to-text serialization suffers from a lack of contextual grounding and representation alignment between the two modalities. To address this gap, we introduce \methodname, a framework that conditions a frozen LLM for nuanced time-series understanding. Instead of direct serialization, \methodname\ enriches the LLM's input with three complementary hints derived from the series: (i) a symbolic numeric hint for numerical grounding, (ii) a context-integrated, step-aligned hint distilled from a pretrained time-series encoder to capture fine-grained dynamics, and (iii) a task-prior hint that encodes global anomaly characteristics. Furthermore, to facilitate robust evaluation of explainability, we introduce a new benchmark featuring multi-format questions and rationales that supervise contextual grounding and pattern-level semantics. Extensive experiments, including both LLM-based and human evaluations, demonstrate that \methodname\ yields explanations of significantly higher quality and achieves competitive detection accuracy compared to general-purpose LLMs, specialized time-series LLMs, and time-series Vision Language Models. 
\footnote{Code is available at \url{https://github.com/thu-sail-lab/AXIS}}.% The code is available in \url{https://anonymous.4open.science/r/TimeSemantic-1742/main.py}
\end{abstract}

\section{Introduction}

\begin{wrapfigure}{r}{0.4\linewidth}
    \centering
    \vspace{-10pt}    
    \includegraphics[width=\linewidth]{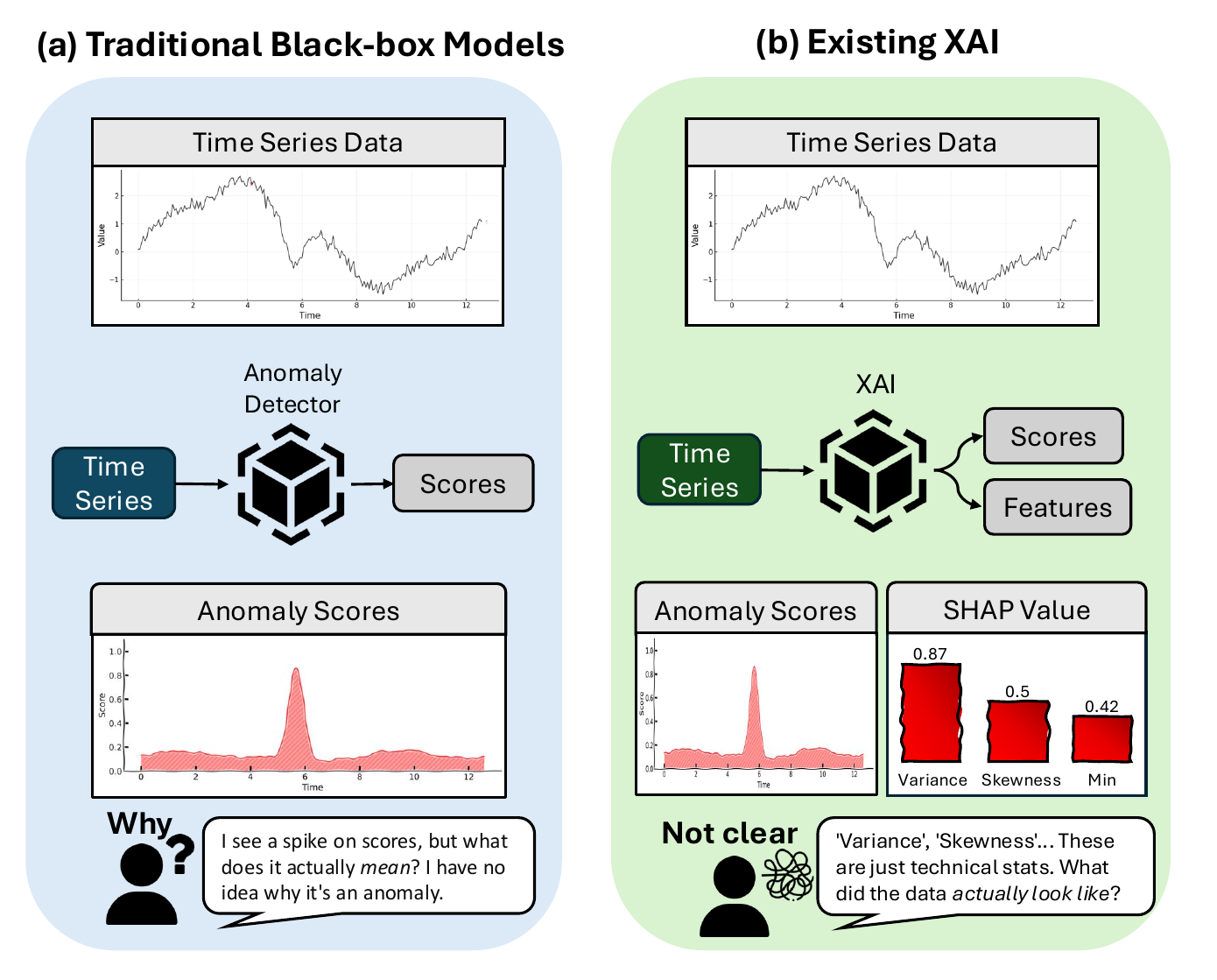}
    \vspace{-15pt}
    \caption{Deep learning method for TSAD: (a) Opaque anomaly scores fail to explain why; (b) XAI features lack intuitive semantics;}
    \vspace{-5pt}
    \label{fig:intro}
\end{wrapfigure}

Time Series Anomaly Detection (TSAD) is essential for safeguarding critical systems across domains~\citep{iqbal2019fault,zeufack2021unsupervised,hundman2018detecting}. While deep learning models can detect anomalies with high accuracy (Fig. \ref{fig:intro}(a)), their adoption in real-world systems is limited by two challenges. First, their reasoning process is essentially a black box. Experts remain in the dark when asking the most practical question: why was this anomaly flagged? Post-hoc attribution methods such as SHAP (Fig.~\ref{fig:intro}(b)) attempt to fill this void, but they merely repackage correlations into statistical features. Such attributions reveal what inputs influenced the model, but they stop short of offering why the underlying anomaly event occurred. Second, these models are brittle. Trained narrowly - often on a single dataset - they capture dataset-specific features rather than generalizable patterns. In fast-changing environments where anomalies manifest in diverse forms, this rigidity is crippling: models must be retrained at high cost, yet still fail to transfer across domains. What is missing is the ability to handle anomalies universally—to recognize and adapt across diverse failure patterns without exhaustive retraining.

\begin{figure}[t]
    \centering
    \includegraphics[width=\linewidth]{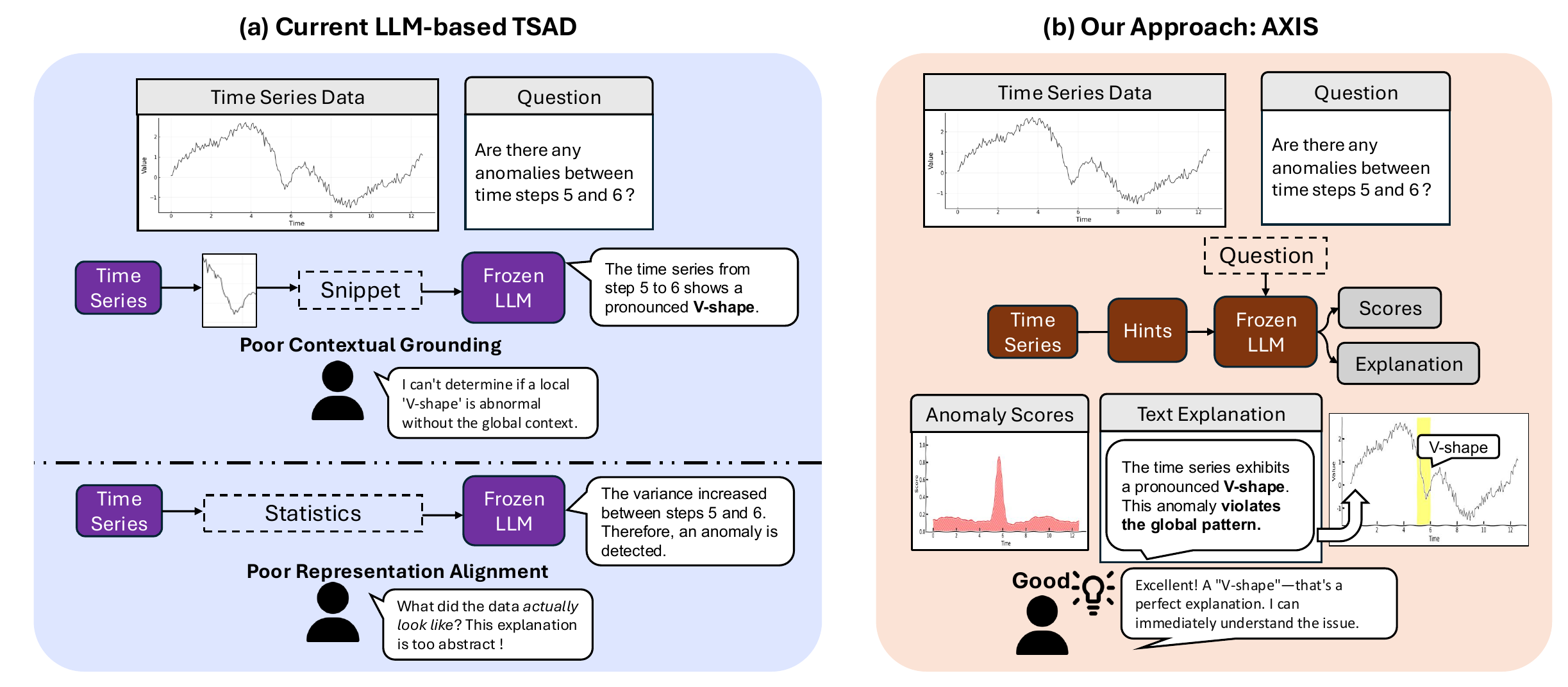}
    \caption{Bridging the Semantic Gap in Time Series Anomaly Explanation. (a) Current LLM-based methods fail due to: (i) poor \textbf{Contextual Grounding}, where observing a local pattern (e.g., the "V-shape") in isolation prevents a meaningful diagnosis; and (ii) \textbf{Representation Misalignment}, where inputs of abstract statistics (e.g., "variance increased") lead to uninformative, circular explanations. (b) Our approach overcomes these limitations by producing contextualized, pattern-level explanations that align with expert reasoning.}
    \label{fig:intro2}
\end{figure}

In response to these limitations, the community has turned to Large Language Models (LLMs), celebrated for their fluency and generalization. Yet critical obstacles remain: LLMs operate on discrete tokens, making them ill-suited for the long, continuous nature of time series ~\citep{dong2024can}. Attempting to fit these signals into tokenized inputs often incurs lossy serialization, forcing workarounds that undercut the LLM’s capabilities. Common strategies—feeding isolated fragments or pre-aggregated statistics—reduce the model to a mere post-hoc translator. As our motivating example illustrates (Fig.~\ref{fig:intro2}(a)), this naive approach, however, suffers a critical semantic gap, driven by two fundamental failures. 

The first is a lack of \textbf{Contextual Grounding}. By analyzing only a narrow snippet of the series, the LLM is deprived of the broader temporal context required to discern whether a local pattern is genuinely anomalous or merely a benign fluctuation. The second is a failure of \textbf{Representation Alignment}, which creates a chasm between the model's analytical basis and human intuition. When an LLM is fed abstract statistical summaries instead of the data's intrinsic shape, its explanations degenerate into shallow echoes of its inputs, failing to provide the qualitative, pattern-level insights that domain experts require to understand what truly happened in the data.

% The first is a failure of \textbf{Contextual Grounding}. When an LLM analyzes only a narrow snippet of the series (e.g., from time step 5 to 6 in Fig.~\ref{fig:intro2}(c)), it is starved of the broader temporal context. The model may correctly recognize a local motif—perhaps labeling it a “V-shape”—but it cannot determine whether this pattern is genuinely anomalous or simply a benign fluctuation within an unseen trend. The result is a description without a diagnosis.
% % 重点标注斜体

% The second is a failure of \textbf{Representation Alignment}. Instead of perceiving the intrinsic shape of the data, the LLM is often fed abstract statistical summaries, such as a sudden variance spike. This creates a semantic chasm between the model's analytic basis and the user's intuitive, pattern-based reasoning. 
% The LLM's output degenerates into shallow echoes of its inputs (“The variance increased... an anomaly is detected” in Fig.~\ref{fig:intro2}(c)), reducing explanation to numbers without insight. Unsurprisingly, domain experts respond with frustration: “That doesn’t tell me what happened. What was the shape of the data?”

Overcoming these failures requires a paradigm shift. Explanations must move beyond statistical paraphrasing toward a native integration of temporal dynamics and linguistic reasoning. This reduces to two core challenges: the \textbf{Contextual Grounding Challenge}, which demands interpreting local events in the context of the full series to explain not only what the data looks like but \textbf{why} it is abnormal; and the \textbf{Representation Alignment Challenge}, which requires bridging the semantic gap between low-level numerical signals and the rich, shape-based concepts underlying human reasoning.

In this paper, we introduce \texttt{\methodname}, a framework designed to address these challenges and unlock the explanatory potential of LLMs for TSAD. Our approach rest on two synergistic contributions. First, to establish the necessary semantic foundation, we construct a novel benchmark with pattern-level labels and rich contextual cues, providing the semantic foundation essential for both grounding and alignment. Second, at the core of our framework is a Hint Tuner that systematically tackles both challenges. For contextual grounding, it distills global time-series information into a compact, informative "hint." For representation alignment, it maps this temporal hint into the LLM's native semantic space. This integrated design transforms a frozen, general-purpose LLM into a  context-aware diagnostic expert, capable of generating correct and high reasonal quality answers for TSAD, as illustrated in Fig.~\ref{fig:intro2}(b). In summary, our main contributions are threefold:
\begin{itemize}[leftmargin=1.5em]
    \item \textbf{A Benchmark for Semantic Explanations:} To bridge the ``semantic gap'' between raw time series signals and linguistic concepts, we construct the first benchmark dedicated to semantic time series anomaly explanation. This benchmark ensures both anomaly diversity and explanation fidelity, providing a principled testbed for evaluating the semantic explainability of TSAD.% It introduces a novel decoupled generation pipeline where a procedural engine synthesizes diverse, pattern-level anomalies, while an LLM generates semantically rich and context-grounded explanations. % 要不要加
    \item \textbf{A Novel Cross-Modal Alignment Framework:} We present \methodname, a framework that aligns a frozen LLM with time-series dynamics. It conditions the LLM on three synergistic inputs: a symbolic numeric hint for numerical grounding, a context-integrated step-aligned hint for fine-grained dynamics, and a task-prior hint for global task priors.
    \item \textbf{Extensive Empirical Validation:} Comprehensive experiments show that \methodname\ establishes a new state of the art in semantic anomaly explanation, substantially outperforming strong baselines including general LLM, specialized time-series LLM, time series VLM. % 修改
\end{itemize}

\section{Related Work}
\label{sec:related}

\paragraph{State-of-the-Art TSAD Models and Interpretability Challenges}
Classical and statistical methods remain competitive baselines yet produce pointwise scores with weak semantics and limited support for multivariate structure~\citep{liu2008isolation,yeh2016matrix}. Deep TSAD improves accuracy via reconstruction~\citep{audibert2020usad,su2019robust,zhang2019deep}, prediction-residual modeling~\citep{tuli2022tranad}, and attention-centric architectures~\citep{xu2021anomaly,yang2023dcdetector,shen2020timeseries,lan2025cicada}, with recent work exploring unified/foundation-style formulations~\citep{shentu2024towards,gao2024units} and diffusion-based detectors~\citep{wang2025diffad}. Explanations, however, are largely post hoc and tied to low-level contributions (e.g., time-step or feature importance), limiting mechanism-oriented diagnosis. Time-series XAI extends attribution~\citep{bento2021timeshap} and investigates prototype/shapelet/motif views and counterfactual recourse~\citep{bahri2022shapelet,yeh2016matrix}, but explanations remain grounded in signal-level statistics rather than pattern-level concepts. This motivates treating TSAD explainability as a semantic alignment problem.

% \subsection{Cross-Modal Representation Learning for Time Series}
% Aligning continuous temporal dynamics with discrete linguistic tokens is a cross-modal challenge. Instruction-style supervision aligns TS with language via synthetic Q\&A or textual augmentation, improving coverage but inheriting generation biases and domain-shift sensitivity~\citep{kong2025time,xie2024chatts}. Contrastive joint embeddings aid retrieval/alignment but often lack conditional generation fidelity without explicit language conditioning~\citep{ito2024clasp}. Parameter-efficient alignment (prompting, adapters, prefix-/p-tuning) injects temporal knowledge into frozen LLMs at low cost~\citep{li2021prefix}. Recent TS–LLM bridges attach numeric/temporal adapters for forecasting or reasoning~\citep{cai2023jolt,chang2025llm4ts,jin2023time,liu2024timecma,liu2024time}, yet typically operate on pre-digested features rather than inducing native, pattern-level semantics. We instead map temporal representations into a soft-prompt space with a time-series soft-prompt tuner (Hint Tuner), conditioning a frozen LLM to internalize pattern-level meaning.

\paragraph{Large Language Models for TSAD}
LLMs can function as zero-shot anomaly detectors under appropriate prompting and input scaling~\citep{alnegheimish2024can,dong2024can,zhou2024can,wang2025chattime}. Performance degrades on long-horizon series due to context-length limits, lossy time-to-text serialization, and chunked inference, which together induce memory decay and boundary artifacts. A complementary line uses LLMs as post-hoc reasoners that verbalize anomaly scores, SHAP attributions, or raw subsequences, or coordinate multi-agent annotation~\citep{liu2025large,lin2024decoding}. In both paradigms, LLMs act mainly as summarizers of low-level signals, yielding descriptive rather than semantically grounded explanations. Our approach directly aligns temporal representations with language via soft-prompt-based conditioning, aiming for faithful, pattern-level explanations.

\paragraph{Benchmarks for TSAD}
Existing benchmarks for time-series question answering, which are adjacent to our task, can be broadly categorized into two paradigms. The first relies on fully synthetic data generation, where normal time series are composed from trends, seasonality, and noise, after which localized anomalies are injected to generate templated or LLM-augmented labels~\citep{cai2024timeseriesexam, xie2024chatts, wang2025chattime, kong2025time2}. The second paradigm uses real-world datasets, pairing authentic time-series data with corresponding semantic information to create evaluation suites~\citep{kim2024multi, cai2025timeseriesgym, liu2024time2, williams2024context, chen2025mtbench}. However, synthetic benchmarks often lack the contextual richness required for robust grounding and representation alignment, while real-world data yields domain-specific explanations that limit model generalizability. To our knowledge, a dedicated benchmark for semantic time series anomaly explanation has remained a critical gap, which our work directly addresses by introducing a benchmark designed for this task. 

\section{Methodology}
This section presents our \methodname\ framework for the semantic anomaly explanation task. 
We begin by formalizing the task in Sec.~\ref{sec:formulation}. 
Next, we introduce the core architecture in Sec.~\ref{sec:methodology}. 
Sec.~\ref{sec:training} describes the two-phase training paradigm—encoder pretraining followed by hint tuning with the LLM frozen—and the inference procedure.
Finally, to enable systematic supervision, we synthesize a benchmark with pattern-level annotations in Sec.~\ref{sec:benchmark}.

\subsection{Problem Formulation} 
\label{sec:formulation}
%We formulate the explainable task of generating natural-language explanations for a target window in a long time series. 

Conventional TSAD methods typically output point-wise anomaly scores for a series of length $T$, but such signals rarely provide human-understandable insights. In practice, anomalies often span contiguous intervals rather than isolated timestamps, and users are chiefly concerned with understanding \textbf{why} an interval is anomalous. To address this, we reformulate the task by introducing a target interval $(s,e)$ and defining the goal as generating a natural-language explanation for it. This window-based formulation respects the temporal continuity of anomalies and makes the explanation task well-posed by localizing reasoning to a specific region within the series. We formalize the problem as follows:

\begin{tcolorbox}[title=Semantic time series anomaly explanation,colback=gray!5, colframe=gray!90]
Given a univariate time series $\mathbf{x}_{1:T}\in\mathbb{R}^{T}$ and a natural-language query $q$, the objective is to explain the pattern within an interval $[s,e)$; in our setup, $(s,e)$ is provided as input. The model learns a mapping $\mathcal{G}$ that, while conditioning on the entire series $\mathbf{x}_{1:T}$ to leverage global context, generates an explanation $\mathbf{y}$ for the target window:
\[
\mathcal{G}: (\mathbf{x}_{1:T}, q, s, e) \mapsto \mathbf{y}.
\]
\end{tcolorbox}
%This window-based formulation is motivated by two considerations. First, in TSAD, anomalies—including point, contextual, and collective types—are localized phenomena with finite temporal support $[s,e)$. Consequently, explanations must be anchored to an explicit window $[s,e)$. Second, it fits within the finite context windows of LLMs by localizing evidence while preserving access to global context.

\subsection{\methodname\ Framework}
\label{sec:methodology}
\begin{figure}[t]
    \centering
    \includegraphics[width=\linewidth]{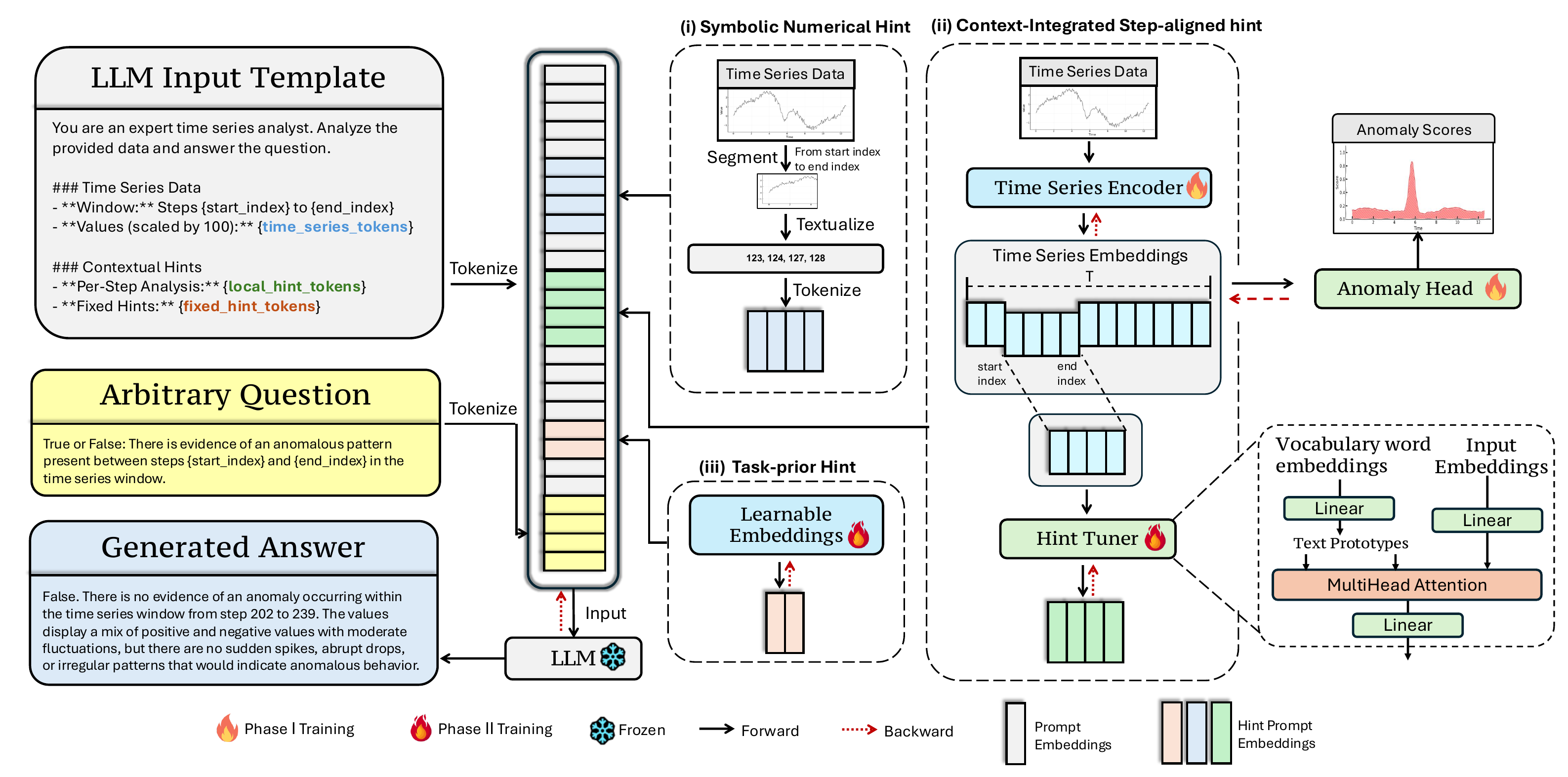}
    \vspace{-5pt}
    \caption{\methodname\ constructs the prompt by three representation pathways: (i) symbolic numeric grounding via window values, (ii) context-integrated local dynamics through step-aligned hints to capture contextual information, and (iii) task-prior hints encoding global priors. \label{fig:framework}} % Time Series Encoder 放在附录里
    \vspace{-10pt}
\end{figure}

We now propose our novel framework called \methodname\  for semantic anomaly explanation task. \methodname\ conditions a frozen LLM through three representation pathways: symbolic numeric hint, context-integrated step-aligned hint, and task-prior hint. The overall framework is shown in Fig.~\ref{fig:framework}. We instantiate this conditioning through three pathways that jointly provide numeric grounding, step-aligned dynamics under global context, and compact task-level priors, without expanding the context length or modifying the LLM.

\paragraph{Symbolic Numeric Hint.}
LLMs possess native reasoning capabilities over discrete numerals when presented symbolically, even in zero-shot settings. To exploit this capability without exhausting the context budget, we textualize only the target window $[s,e)$ after Z-score normalization of the full series $\mathbf{x}_{1:T}$. Values are scaled by a factor $r$ (default $r{=}100$) to preserve precision while avoiding decimal tokens~\citep{liu2024time}, rounded to integers, and serialized as a delimiter-separated string (e.g., ``123, 124, 127, 128'') to constitute \texttt{\{time\_series\_tokens\}}.
This pathway is compact—its position cost scales as $\alpha (e-s)+c$ where $\alpha$ is the average subword tokens per integer and $c$ the delimiter overhead—yet it preserves step-wise numeric grounding.

\paragraph{Context-Integrated Step-aligned Hint.}

% \begin{wrapfigure}{r}{0.5\linewidth}
%     \centering
%     \vspace{-10pt}    
%     \includegraphics[width=\linewidth]{figs/HintTuner.png}
%     \vspace{-15pt}
%     \caption{Structure of the Hint Tuner.}
%     \label{fig:Encoder-Tuner}
%     \vspace{-7.5pt}
% \end{wrapfigure}
While the above textualization provides direct numeric access, it cannot capture long-range dependencies essential for TSAD such as regime shifts, seasonality interactions, and boundary effects. We therefore condense global information into step-aligned local representations via a pretrained time-series encoder and a \emph{Hint Tuner}, in the spirit of \citep{jin2023time}. A Transformer encoder $f_{\theta}$ consumes $\mathbf{x}_{1:T}$ and outputs embeddings $\mathbf{H}_{1:T}\in\mathbb{R}^{T\times d_{\text{proj}}}$ where $d_{\text{proj}}$ is the projection dimension (the details are shown in Appx.~\ref{appx:ts-encoder}). We slice $\mathbf{H}_{s:e}$ and map it into the LLM space using cross-attention over a prototype bank derived from the LLM vocabulary: $\mathbf{S}_{\text{proto}}=\mathbf{M}\mathbf{E}_{\text{vocab}}$ where $\mathbf{E}_{\text{vocab}}\in\mathbb{R}^{|\mathcal{V}|\times d_h}$ is fixed word embeddings and $\mathbf{M}\in\mathbb{R}^{P\times |\mathcal{V}|}$ is a learnable linear projection. Queries $\mathbf{W}_q\,\mathbf{H}_{s:e}$ attend to this bank:
$
\tilde{\mathbf{H}}_{s:e} = \mathrm{Attn}\!\big(\mathbf{W}_q\, \mathbf{H}_{s:e},\, \mathbf{S}_{\text{proto}},\, \mathbf{S}_{\text{proto}}\big)\ \in\ \mathbb{R}^{(e-s)\times d_h}.
$
The resulting $\tilde{\mathbf{H}}_{s:e}$ acts as step-aligned local hints that inject global context and temporal structure into the LLM embedding space while keeping both the LLM and $f_{\theta}$ frozen. This pathway adds $(e-s)$ positions linearly while supplying detailed global context and temporal alignment. %【解释一下作用】

\paragraph{Task-Prior Hint.}
To regularize decoding and inject task-level priors that remain stable across instances, we introduce a small set of shared queries $\mathbf{P}_{\text{fix}}\in\mathbb{R}^{K\times d_h}$ that attend to the same prototype source:
$
\tilde{\mathbf{F}} = \mathrm{Attn}\!\big(\mathbf{P}_\text{fix},\, \mathbf{S}_{\text{proto}},\, \mathbf{S}_{\text{proto}}\big)\ \in\ \mathbb{R}^{K \times d_h}.
$

\paragraph{Final Prompt.} The three hint pathways are integrated into a unified prompt that conditions the frozen LLM, as illustrated in Fig.~\ref{fig:framework}. The final input sequence is constructed from a template containing the user's query $q$, the textualized window values, and special placeholder tokens. At input time, the embeddings for the $K$ task-prior hints ($\tilde{\mathbf{F}}$) and the $(e-s)$ step-aligned hints ($\tilde{\mathbf{H}}_{s:e}$) replace the embeddings of $K+(e-s)$ placeholder tokens. The symbolic numeric hint is inserted directly as text. This process yields a single, coherent input sequence for the LLM that combines natural language with rich, multi-faceted temporal information, all without requiring architectural changes to the base model.

\subsection{Training Objective and Inference}
\label{sec:training}

Our method is trained in two phases. First, we pretrain the time-series encoder $f_{\theta}$ with a joint objective of masked reconstruction and anomaly classification with an additional linear head, then freeze its weights. Second, keeping both the encoder and the LLM frozen, we train only the Hint Tuner and its associated parameters ($\mathbf{M}$, $\mathbf{P}_{\text{fix}}$). This phase optimizes a next-token prediction loss to generate the explanation $\mathbf{y}$ conditioned on the query and all three hint pathways. The detailed training process is given in Appx.~\ref{appx:training}.

\subsection{A Benchmark for semantic time series anomaly explanation}
\label{sec:benchmark}

Existing methods reveal a foundational limitation: the community lacks a benchmark that teaches models to speak the language of temporal patterns. To address both the contextual grounding and representation alignment challenges outlined earlier, we synthesize a benchmark specifically designed to train models to reason about anomalies like human experts. Rather than a mere collection of time series, our benchmark constitutes a carefully curated curriculum built around three core design principles.

\begin{wrapfigure}{r}{0.5\linewidth}
    \centering
    \vspace{-10pt}    
    \includegraphics[width=\linewidth]{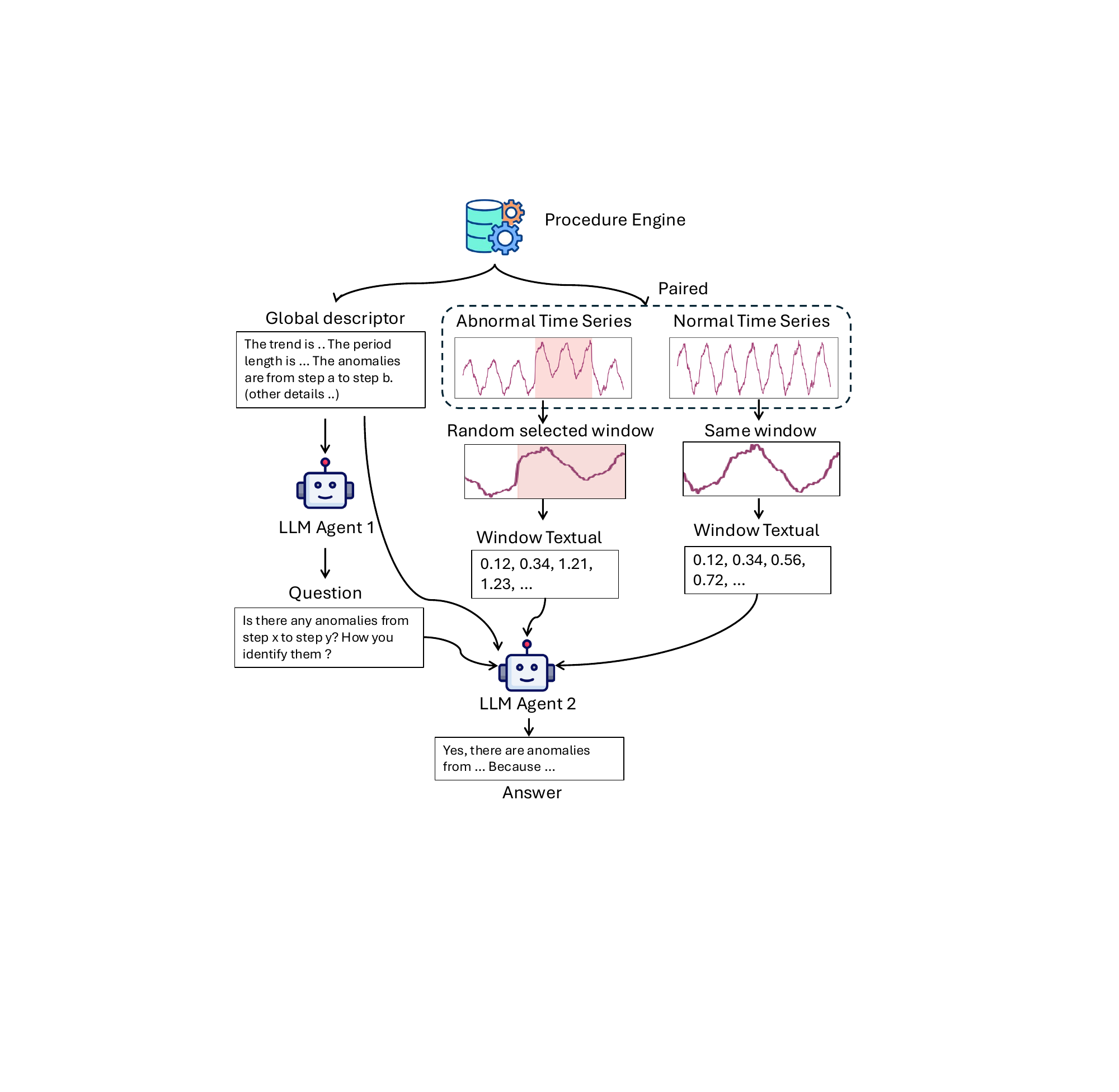}
    \vspace{-15pt}
    \caption{The architecture of our procedural engine for generating context-aware and comparative anomaly explanation benchmarks.}
    \label{fig:engine}
    \vspace{-10pt}
\end{wrapfigure}

\paragraph{Pattern-Level Anomaly Vocabulary.} To address the representation alignment challenge, we introduce a procedural engine that moves beyond abstract statistical deviations to a vocabulary of interpretable, pattern-level anomalies using the procedure of \citet{lan2025towards}. As illustrated in Figure~\ref{fig:engine}, our engine synthetically composes canonical anomaly primitives—such as \textit{sudden spikes}, \textit{level shifts}, and \textit{periodicity breaks}—onto clean baseline series. A key advantage of our approach is the generation of \textbf{paired time series}: for every abnormal series created, a corresponding normal counterpart is preserved. This methodology establishes an unambiguous, verifiable link between anomaly time series and its linguistic label, forming the bedrock for teaching models to reason about the semantics of temporal events.

\paragraph{Contextual and Comparative Reasoning.}
To overcome the contextual grounding challenge, we designed our benchmark to compel models to reason about local events within a global and comparative framework. Naively presenting isolated time-series windows is insufficient. Instead, our engine first generates a \textbf{global descriptor}, a textual summary of the series' overall dynamics (e.g., trends, seasonality), which provides essential context. Second, we employ a comparative windowing strategy. A model is presented not only with a window containing a potential anomaly but also with the corresponding temporal window from its “healthy” paired series. This core design choice is a significant advantage, as it inherently frames the task as a discriminative one: the model must learn to articulate \textbf{why} a specific pattern deviates from an explicit, provided norm, rather than merely describing a segment in isolation.

\paragraph{LLM-Powered Explanation Generation.}
Building on this structured foundation, we leverage LLMs to generate high-quality, multi-format supervision signals. As depicted in our pipeline, this is a multi-agent process. One LLM agent uses the global descriptor to formulate a targeted diagnostic question. A second, more powerful agent is then tasked with answering this question, conditioned on the global descriptor, the abnormal and normal window data. The primary motivation here is to generate rich, conceptual explanations. Our prompts are meticulously engineered to discourage superficial strategies (e.g., quoting raw values) and instead elicit reasoning based on the intrinsic, morphological characteristics of the anomaly. This process yields a diverse and consistent set of questions and detailed rationales, creating a powerful supervisory signal for training.

\paragraph{Ensuring Benchmark Integrity.}
To guarantee the scientific utility and integrity of our benchmark, we implement a rigorous quality control pipeline. This process verifies the agreement between ground-truth labels and generated answers, enforces stylistic consistency, and filters potential redundancies. We provide comprehensive dataset statistics and are releasing all generation metadata to ensure the full reproducibility of our benchmark. The result is not merely a dataset but a robust training environment engineered to finally bridge the semantic gap in time series anomaly explanation. Some examples of the generated Q\&A pairs are given in Appx.~\ref{appx:qa-exemplars}.

\section{Experiments}
\label{sec:experiments}

To validate the effectiveness of \methodname, we conduct a series of experiments designed to answer three central research questions: \textbf{RQ1: Explanation Quality.} How does \methodname\ compare against state-of-the-art LLM-based methods in generating high-quality, semantic anomaly explanations? \textbf{RQ2: Component Importance.} How do the core components of our framework—the symbolic numeric hint, the context-integrated step-aligned hint, and the task-prior hint—contribute to the final explanation quality? \textbf{RQ3: Architectural Generality.} How robust is the \methodname\ framework when applied to different underlying frozen LLMs? Finally, in Appx.~\ref{appx:training}, we demonstrate that our Phase I TSAD model achieves results comparable to state-of-the-art methods on real-world public TSAD datasets.
\subsection{Experimental Setup}

\paragraph{Dataset.}
All experiments are conducted on our newly created \textbf{Semantic Anomaly Benchmark} (detailed in Sec.~\ref{sec:benchmark}). This benchmark is specifically designed for the task of semantic time series anomaly explanation, containing diverse anomaly patterns, multi-format questions (Multiple Choice, True/False, Open-Ended), and detailed, pattern-level ground-truth explanations. The all hyperparameters for \methodname\ is given in Appx. \ref{appx:hyperparameters}.

\paragraph{Baselines.}
We compare \methodname\ with a comprehensive set of strong baselines, categorized as follows:
\textbf{Timeseries VLM:}\citep{he2025harnessing} \texttt{Image LLM} is supported by gpt-4o, which analyzes plots of the full time series with highlighted window, treating the explanation task as a visual reasoning problem. \textbf{Specialized TS-LLM Methods:} We include several recent models designed for time series analysis with LLMs: \texttt{ChatTS}~\citep{xie2024chatts}, \texttt{LLMAD}~\citep{liu2025large}, \texttt{ChatTime}~\citep{wang2025chattime}, and \texttt{AnomLLM}~\citep{dong2024can}. We evaluate \texttt{AnomLLM} in two settings: providing the full series (\texttt{AnomLLM(Full)}) and providing only the target window (\texttt{AnomLLM(Window)}).

\paragraph{Evaluation Metrics.}
Following recent work on evaluating LLM-generated content, we use an LLM-as-a-judge approach, specifically \textbf{G-eval} \citep{liu2023g} with Gemini-2.5 as the arbiter. The quality of explanations is assessed across multiple dimensions tailored to each question type, including Correctness (Corr.), Reasoning Quality (Rsn. Qual.), Accuracy (Acc.), Completeness (Comp.), Relevance (Rel.), and Justification Quality (Justif.). A final, holistic score (Final) is also computed. The detailed definition for evaluation metrics is given in Appx.~\ref{appx:g-eval}

\subsection{Main Results: Explanation Quality (RQ1)}

Table~\ref{tab:main-table} presents the main results comparing our model, \methodname, against all baselines. Our method demonstrates superior performance across all metrics and question types, establishing a new state-of-the-art for the task.

Specifically, \methodname\ achieves the highest final scores on Multiple Choice (4.19), Open-Ended (3.02), and True/False (3.65) questions. This consistent top-ranking performance highlights its robust ability to generate accurate, complete, and well-reasoned explanations regardless of the question format. Compared to specialized TS-LLM baselines like \texttt{ChatTS} and the \texttt{AnomLLM} variants, our method shows a significant improvement, underscoring the effectiveness of our proposed hint-based conditioning strategy. The strong performance against the \texttt{Image LLM} baseline further suggests that our multi-pathway representation provides richer, more aligned signals for the LLM than raw visual serialization.

\begin{table*}[t]
\centering
\small
\caption{Main Results: \methodname\ vs Baselines}
\resizebox{\textwidth}{!}{%
\begin{tabular}{l||ccc|cccc|ccc}
\hline
 & \multicolumn{3}{c}{Multiple Choice} & \multicolumn{4}{c}{Open Ended} & \multicolumn{3}{c}{True False} \\
Model & Final & Corr. & Rsn. Qual. & Final & Acc. & Comp. & Rel. & Final & Corr. & Justif. \\
\hline
\textbf{\methodname} & \textbf{4.19} & \textbf{4.21} & \textbf{4.14} & \textbf{3.02} & \textbf{2.87} & \textbf{2.93} & \textbf{3.31} & \textbf{3.65} & \textbf{3.60} & \textbf{3.74} \\
Image LLM & \underline{4.09} & \underline{4.12} & \underline{4.02} & 2.68 & 2.53 & 2.49 & 3.07 & 2.64 & 2.57 & 2.74 \\
ChatTS & 3.29 & 3.40 & 3.05 & 2.19 & 1.67 & 2.13 & 2.87 & 2.79 & 2.76 & 2.83 \\
LLMAD & 2.73 & 2.70 & 2.79 & 2.09 & 2.09 & 1.89 & 2.31 & 2.49 & 2.52 & 2.43 \\
ChatTime & 1.33 & 1.49 & 0.98 & 0.96 & 0.95 & 0.98 & 0.95 & 1.04 & 1.07 & 1.00 \\
AnomLLM(Full) & 3.13 & 2.98 & 3.49 & \underline{2.86} & 2.53 & \underline{2.89} & 3.20 & 2.88 & 2.60 & \underline{3.31} \\
AnomLLM(Window) & 3.78 & 3.81 & 3.70 & 2.84 & \underline{2.78} & 2.55 & \underline{3.24} & \underline{3.32} & \underline{3.45} & 3.12 \\
\hline
\end{tabular}
}%
\vspace{-2.5pt}
\label{tab:main-table}
\end{table*}

\begin{figure}[t]
    \centering
    \includegraphics[width=\linewidth]{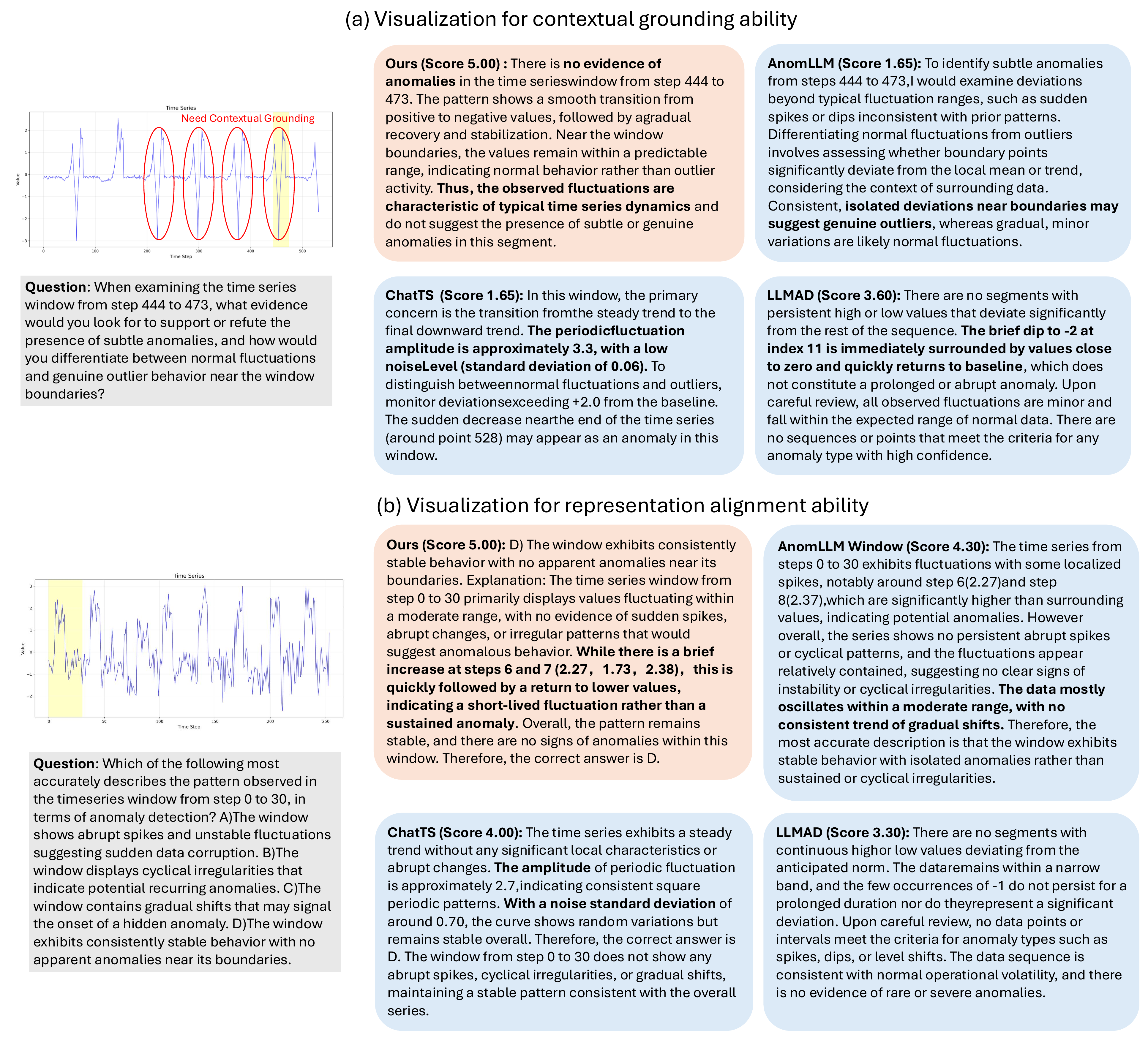}
    \vspace{-15pt}
    \caption{Visualization of (a) contextual grounding and (b) representation alignment ability}
    \label{fig:visualization}
    \vspace{-10pt}
\end{figure}

\paragraph{Visualization.} To qualitatively illustrate these performance gains, Fig.~\ref{fig:visualization} presents a comparative case study. In the Fig.~\ref{fig:visualization}(a), the target window (steps 444 to 473) exhibits pronounced oscillations. \methodname\ correctly contextualizes these dynamics against the broader series, identifying them as part of a normal periodic pattern and concluding there is no anomaly. In stark contrast, a baseline like AnomLLM or ChatTS, when limited to the window view, lacks this broader context and erroneously flags the internal deviations as potential outliers. In Fig.~\ref{fig:visualization}(b), \methodname\ provides a precise characterization by explicitly identifying the brief increase at steps 6–7 (2.27, 1.73, 2.38) and correctly interpreting it as a transient fluctuation rather than a sustained anomaly, concluding that the pattern is stable and anomaly-free. In contrast, alternative approaches tend to give vague or generalized descriptions, often noting fluctuations or moderate oscillations without distinguishing whether they indicate normal behavior or anomalies. This comparison highlights that our representation alignment framework enables fine-grained, context-aware interpretation of time series behavior, avoiding ambiguous assessments and ensuring robust anomaly detection.

\begin{figure}[t]
    \centering
    \includegraphics[width=\linewidth]{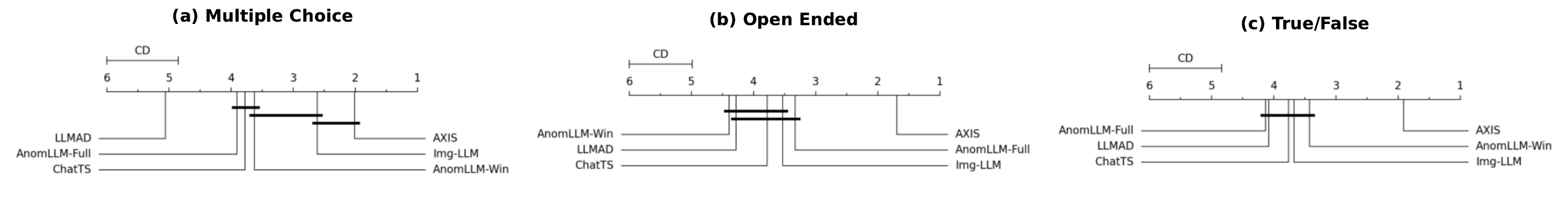}
    \vspace{-15pt}
    \caption{Critical Difference diagrams illustrating the statistical comparison of model performance based on human rankings for (a) Multiple Choice, (b) Open-Ended, and (c) True/False questions.}
    \label{fig:human-eval}
    \vspace{-10pt}
\end{figure}

\paragraph{Human evaluation.} To further validate these quantitative results from a human-centric perspective, we also performed a statistical analysis based on expert rankings of the model-generated explanations. We conducted a survey where human evaluators were asked to rank the outputs from all competing models for each question type. Fig.~\ref{fig:human-eval} corroborates the findings from Table~\ref{tab:main-table}(We remove the ChatTime for poor performance). Our model, \methodname, consistently achieves the best (lowest) average rank across all three tasks, positioning it furthest to the right. The details for experiment setup and results are shown in Appx.~\ref{appx:human}.

\begin{table*}[t]
\centering
\small
\caption{Ablation Studies of Hint Components \label{tab:ablation}}
\resizebox{\textwidth}{!}{%
\begin{tabular}{l||ccc|cccc|ccc}
\hline
 & \multicolumn{3}{c}{Multiple Choice} & \multicolumn{4}{c}{Open Ended} & \multicolumn{3}{c}{True False} \\
Model & Final & Corr. & Rsn. Qual. & Final & Acc. & Comp. & Rel. & Final & Corr. & Justif. \\
\hline
\methodname\ & \textbf{4.19} & \textbf{4.21} & \textbf{4.14} & \textbf{3.02} & \textbf{2.87} & \textbf{2.93} & \textbf{3.31} & \textbf{3.65} & \textbf{3.60} & \textbf{3.74} \\
w/o-task-hint & 3.82 & 3.93 & 3.56 & 2.33 & 2.13 & 2.22 & 2.69 & \underline{3.25} & \underline{3.31} & \underline{3.17} \\
w/o-context-hint & \underline{4.09} & \underline{4.16} & \underline{3.91} & \underline{2.75} & \underline{2.56} & \underline{2.58} & \underline{3.16} & 2.44 & 2.48 & 2.38 \\
w/o-windows & 3.95 & 4.00 & 3.84 & 2.41 & 2.00 & 2.36 & 2.95 & 2.87 & 2.83 & 2.93 \\
\hline
\end{tabular}
}%
\end{table*}

\begin{table*}[t]
\centering
\small
\caption{\methodname\ variants across LLM families and settings (standardized naming: family + size + variant; \textit{Instruct} denotes instruction-tuned, \textit{Coder} denotes code-pretrained}
\label{tab:variants}
\resizebox{\textwidth}{!}{%
\begin{tabular}{l l||ccc|cccc|ccc}
\hline
 &  & \multicolumn{3}{c}{Multiple Choice} & \multicolumn{4}{c}{Open Ended} & \multicolumn{3}{c}{True False} \\
Family & Variant & Final & Corr. & Rsn. Qual. & Final & Acc. & Comp. & Rel. & Final & Corr. & Justif. \\
\hline
Deepseek-Llama & 8B (Instruct) & 4.28 & 4.30 & 4.23 & 3.02 & \underline{2.84} & 2.84 & \underline{3.45} & 3.64 & 3.55 & \textbf{3.79} \\
Deepseek-Qwen & 14B (Instruct) & \underline{4.31} & 4.28 & \underline{4.37} & \underline{3.03} & 2.80 & \underline{2.93} & 3.42 & 3.60 & 3.55 & 3.69 \\
Deepseek-Qwen & 7B (Instruct) & 4.19 & 4.21 & 4.14 & 3.02 & \textbf{2.87} & \underline{2.93} & 3.31 & \underline{3.65} & \underline{3.60} & \underline{3.74} \\
Deepseek-Qwen & 1.5B (Instruct) & 4.07 & 4.12 & 3.95 & 2.72 & 2.65 & 2.55 & 3.00 & 3.18 & 3.17 & 3.19 \\
Qwen2.5 & 7B (Coder) & \textbf{4.40} & \textbf{4.40} & \textbf{4.42} & 2.72 & 2.64 & 2.58 & 2.98 & 2.96 & 2.98 & 2.93 \\
Qwen2.5 & 7B (Base) & 4.30 & \underline{4.37} & 4.12 & 2.75 & 2.73 & 2.45 & 3.13 & \textbf{3.66} & \textbf{3.69} & 3.62 \\
Qwen2.5 & 7B (Instruct) & 4.17 & 4.22 & 4.06 & \textbf{3.08} & 2.80 & \textbf{3.00} & \textbf{3.50} & 3.11 & 3.00 & 3.27 \\
Mistral & 7B (Base) & 2.97 & 3.05 & 2.79 & 2.89 & 2.69 & 2.82 & 3.20 & 2.69 & 2.55 & 2.90 \\
Mistral & 7B (Instruct) & 3.36 & 3.33 & 3.44 & 2.77 & 2.58 & 2.45 & 3.35 & 3.27 & 3.17 & 3.43 \\
\hline
\end{tabular}
}%
\end{table*}

\subsection{Ablation Studies (RQ2)}

To understand the contribution of each component in \methodname, we conducted an ablation study, with results summarized in Table~\ref{tab:ablation}. The study confirms that all three input pathways are crucial for optimal performance. Removing the context-integrated step-aligned hint (`w/o-context-hint`) degrades performance most severely on True/False questions (Final score drops from 3.65 to 2.44), confirming that binary judgments rely on precise, step-aligned temporal cues. In contrast, Open-Ended questions are most sensitive to the symbolic numeric hint from the window (`w/o-windows`); without it, accuracy drops markedly (2.00 vs 2.87) and overall quality declines (Final 2.41 vs 3.02), indicating that direct numeric access provides the fine-grained grounding needed for detailed answers. Eliminating the task-prior hint (`w/o-task-hint`) also substantially harms Open-Ended completeness and relevance (2.22/2.69 vs 2.93/3.31), suggesting that these global priors help structure the explanation and ensure comprehensive coverage. The full model consistently yields the best scores, validating our design choices. Additional experiments for causal contribution of hints are given in Appx.~\ref{appx:causal-ablation}.

\subsection{Analysis of Architectural Variants (RQ3)}

To assess its generality, we instantiated \methodname\ across multiple LLM families and settings, using a standardized \emph{Family + Size + Variant} naming scheme and a fixed data schedule (R1) to isolate architectural effects. As shown in Table~\ref{tab:variants}, our framework demonstrates robust cross-family adaptation (Qwen, Llama, Mistral) and its performance scales with model size (e.g., Qwen-14B $>$ 7B $>$ 1.5B). The results also reveal complementary strengths among model variants: code-pretrained models like \textit{Qwen2.5-7B Coder} excel at structured discrimination tasks, achieving the highest Multiple Choice score (4.40), whereas instruction-tuned versions such as \textit{Qwen2.5-7B Instruct} lead in free-form explanatory quality, with top scores in Open-Ended relevance and completeness (3.50/3.00). This highlights that \methodname\ not only universally enhances different base models but also allows for trade-offs between discriminative and explanatory objectives through strategic variant selection.

\section{Conclusion}
\label{sec:conclusion}
We introduce a novel cross-modal framework that effectively adapts frozen Large Language Models for semantic time series anomaly explanation. By using a three-stream conditioning strategy that combines a symbolic numeric hint, a context-integrated step-aligned hint, and a task-prior hint, our method achieves strong performance in both detection accuracy and explanation quality. Future work will explore incorporating domain-specific knowledge graphs to enhance causal reasoning and generating multi-modal explanations that include visualizations alongside text.

\section*{Ethics statement}

This work focuses on explainable time series anomaly detection and does not involve personally identifiable information or other sensitive attributes. Our benchmark primarily uses procedurally synthesized data and publicly available datasets; no private logs are collected, and no re-identification is attempted. Human evaluation was conducted with informed consent, anonymized responses, and fair compensation in line with institutional guidelines, without storing any personally identifying information. 

\section*{Reproducity statement}

We aim for full reproducibility. Upon publication, we will release code, configuration files, and scripts to reproduce: (i) the benchmark synthesis pipeline (including prompts, fixed random seeds, and parameter settings); (ii) Phase I encoder pretraining and Phase II hint tuning with exact hyperparameters, token budgets, and training schedules (see Appendix~\ref{appx:training} and Appendix~\ref{appx:hyperparameters}); and (iii) evaluation pipelines, including baseline configurations, G-Eval judge prompts, scoring scripts, and human-evaluation materials. We will provide model checkpoints where licensing permits or otherwise specify exact model identifiers and initialization procedures. % The code is available in \url{https://anonymous.4open.science/r/TimeSemantic-1742/main.py}

\bibliography{iclr2026_conference}
\bibliographystyle{iclr2026_conference}

\newpage
\appendix

\definecolor{codebg}{RGB}{250,250,250}
\definecolor{coderule}{RGB}{220,220,220}
\definecolor{codetext}{RGB}{20,20,20}
\lstdefinestyle{prompt}{
  basicstyle=\ttfamily\small\color{codetext},
  frame=single,
  rulecolor=\color{coderule},
  backgroundcolor=\color{codebg},
  breaklines=true,
  columns=fullflexible,
  showstringspaces=false,
  upquote=true,
  aboveskip=6pt,
  belowskip=6pt,
  xleftmargin=2pt,
  framexleftmargin=2pt
}
\lstdefinestyle{qa}{
  basicstyle=\ttfamily\small\color{codetext},
  frame=single,
  rulecolor=\color{coderule},
  backgroundcolor=\color{codebg},
  breaklines=true,
  columns=fullflexible,
  showstringspaces=false,
  upquote=true,
  aboveskip=6pt,
  belowskip=6pt,
  xleftmargin=2pt,
  framexleftmargin=2pt
}

\section{Use of LLMs}
Our manuscript preparation involved the use of large language models (LLMs) primarily for refining language, improving grammar, readability, and stylistic elements. Crucially, LLMs also constituted a core component of our research process. They were leveraged as agents to participate in data generation, established as benchmarks for experimental baselines, and contributed to the comprehensive evaluation of experimental results. For a detailed account of LLMs' specific applications within our methodology, please refer to Section ~\ref{sec:benchmark} and Section ~\ref{sec:experiments}. We confirm that the contributions of LLMs, whether in writing or research, do not impede the reproducibility of our reported findings.

\section{Details for \methodname\ }
\subsection{Time-Series Encoder Details}\label{appx:ts-encoder}

\begin{wrapfigure}{r}{0.3\linewidth}
    \centering
    \vspace{-10pt}    
    \includegraphics[width=\linewidth]{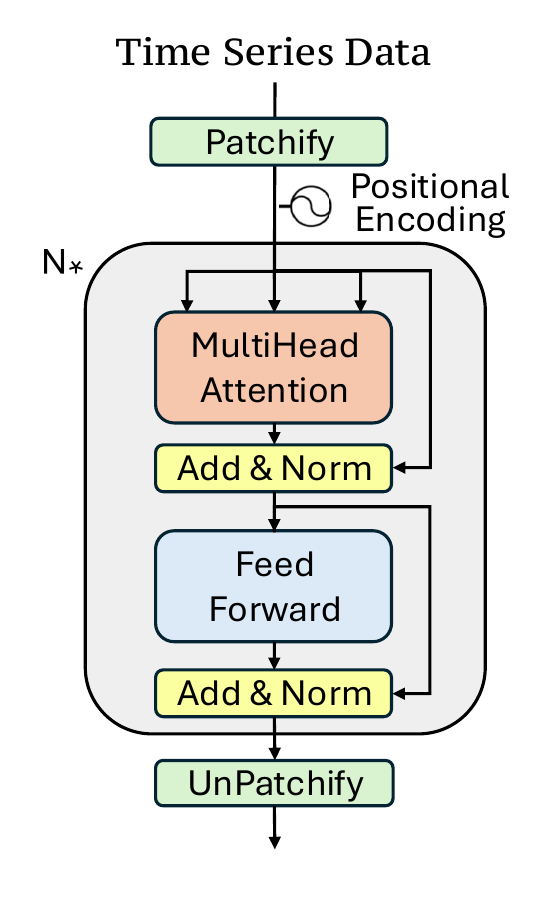}
    \vspace{-18pt}
    \caption{The structure of time-series encoder.}
    \label{fig:timeseries-encoder}
    \vspace{-5pt}
\end{wrapfigure}

This section details the architecture of the time-series encoder $f_{\theta}$ used in \methodname. The overall structure is shown in Fig.~\ref{fig:timeseries-encoder}.

\paragraph{Patchify.} Let $P$ denote the patch size. We pad $\mathbf{x}_{1:T}$ to length $T' = \lceil T/P \rceil P$ with zeros and form $N = T'/P$ contiguous, non-overlapping patches:
\[
\mathbf{X}^{(n)} = [x_{(n-1)P+1},\dots,x_{nP}]\in\mathbb{R}^{P},\quad n=1,\dots,N.
\]
Stacking yields $\mathbf{X}_p\in\mathbb{R}^{N\times P}$. The patch-level attention mask is obtained by aggregating the step mask: a patch is valid if it contains at least one valid step,
\[
\mathbf{m}^{\text{patch}}_n = \mathbb{I}\Big(\sum_{t=(n-1)P+1}^{nP} m_t > 0\Big),\quad n=1,\dots,N.
\]

\paragraph{Patch Embedding.} Each patch is projected to the model dimension $d_{\text{model}}$ with a linear layer
\[
\mathbf{Z} = \mathbf{X}_p \mathbf{W}_e + \mathbf{1}\,\mathbf{b}_e^{\top} \in \mathbb{R}^{N\times d_{\text{model}}},\quad \mathbf{W}_e\in\mathbb{R}^{P\times d_{\text{model}}}.
\]

\paragraph{Positional Encoding via RoPE.} The encoder uses rotary positional encoding (RoPE) applied \emph{inside} self-attention to the query and key of each head. Concretely, for a sequence of $N$ patch tokens, RoPE generates frequency pairs that rotate the $d_{\text{model}}$ channels in 2D subspaces, providing translation-friendly relative position information.

\paragraph{Transformer Encoder (Non-causal).} We stack $L$ pre-norm Transformer encoder layers. Each layer comprises: (i) multi-head self-attention with $H$ heads and head dimension $d_{\text{head}}=d_{\text{model}}/H$; (ii) an MLP block of the LLaMA style with gated GELU activation. RMSNorm is used before attention and MLP. Attention is \emph{non-causal} and operates over all $N$ patches, enabling global context aggregation. The attention mask derived from $\mathbf{m}^{\text{patch}}$ prevents padded tokens from contributing. In the univariate case there is no inter-feature bias term; the attention is standard scaled dot-product with RoPE.

\paragraph{UnPatchify and Projection to Step Alignment.} The encoder output $\mathbf{U}\in\mathbb{R}^{N\times d_{\text{model}}}$ is mapped to a patch-level representation of size $P\times d_{\text{proj}}$ through a linear projection
\[
\mathbf{Y} = \mathbf{U} \mathbf{W}_p \in \mathbb{R}^{N\times (P\, d_{\text{proj}})},\quad \mathbf{W}_p\in\mathbb{R}^{d_{\text{model}}\times (P\, d_{\text{proj}})}.
\]
Reshaping and reordering yield step-aligned embeddings $\tilde{\mathbf{H}}\in\mathbb{R}^{T'\times d_{\text{proj}}}$. We finally drop the padded tail to obtain
\[
\mathbf{H}_{1:T} = \tilde{\mathbf{H}}_{1:T} \in \mathbb{R}^{T\times d_{\text{proj}}}.
\]

\subsection{Training Process Details}
\label{appx:training}

Our model is trained in a two-phase process designed to first build a robust time-series representation and then align it with the frozen LLM's semantic space.
% \chen{maybe use PHASE I, II instead of STAGE}
\subsubsection{Phase I: Time-Series Encoder Pretraining}
In the first phase, we pretrain the time-series encoder $f_{\theta}$ to learn a versatile representation of temporal dynamics. This is   achieved by optimizing a joint objective function that combines masked reconstruction and anomaly classification. The encoder is a Transformer-based architecture that processes the entire input time series $\mathbf{x}_{1:T}$.

\paragraph{Masked Reconstruction.} Following the principles of self-supervised learning for sequential data, we employ a masked reconstruction objective. A portion of the input time series is randomly masked, and the encoder is tasked with reconstructing the masked values. Let $\mathbf{x}$ be the input series and $\mathbf{m}$ be a binary mask where $m_i=1$ if the $i$-th timestep is masked and $0$ otherwise. The encoder $f_{\theta}$, followed by a reconstruction head $g_{\psi}$, outputs a reconstructed series
\[
\hat{\mathbf{x}} = g_{\psi}\big(f_{\theta}(\mathbf{x} \odot (\mathbf{1}-\mathbf{m}))\big).
\]
The reconstruction loss is the Mean Squared Error (MSE) over the masked timesteps:
\[
\mathcal{L}_{\text{recon}} = \frac{1}{\sum_i m_i} \sum_{i=1}^{T} m_i (x_i - \hat{x}_i)^2.
\]
This objective forces the encoder to learn the underlying patterns and dependencies within the time series.

\paragraph{Anomaly Classification.} To explicitly teach the encoder to distinguish between normal and anomalous patterns. Given a time series $\mathbf{x}$ and its corresponding binary anomaly label $\mathbf{l} \in \{0, 1\}^T$, the model predicts the probability of an anomaly $\hat{\mathbf{l}}_t=h_{\zeta}(f_\theta(\mathbf{x}))$. We use the BCE loss:
\[
\mathcal{L}_{\text{class}} = - \sum_{t=1}^T\big( l_t \log(\hat{l}_t) + (1-l_t)\log(1-\hat{l}_t) \big).
\]

\paragraph{Joint Objective.} The total loss for the pretraining phase is a weighted sum of the reconstruction and classification losses:
\[
\mathcal{L}_{\text{pretrain}} = \lambda_{\text{recon}} \mathcal{L}_{\text{recon}} + \lambda_{\text{class}} \mathcal{L}_{\text{class}},
\]
where $\lambda_{\text{recon}}$ and $\lambda_{\text{class}}$ are hyperparameters that balance the two objectives. After this phase, the weights of the encoder $f_{\theta}$ are frozen (only $f_{\theta}$ is retained for the next phase).

\subsubsection{Phase II: Hint Tuner Training}
In the second phase, we focus on aligning the pretrained time-series representations with the LLM. Both the time-series encoder $f_{\theta}$ and the LLM are kept frozen to maintain their powerful, pre-existing capabilities. The only trainable components are the Hint Tuner and its associated parameters, namely the prototype bank selection matrix $\mathbf{M}$ and the learnable task-prior queries $\mathbf{P}_{\text{fix}}$.

The goal of this phase is to train the model to generate a natural language explanation $\mathbf{y} = (y_1, y_2, \dots, y_N)$ conditioned on the time series $\mathbf{x}$, the query $q$, and the derived hints. The training objective is a standard causal language modeling loss, which maximizes the likelihood of the ground-truth explanation.

Specifically, the input to the LLM is constructed by embedding the query $q$, the symbolic-numeric hint, and replacing placeholder token embeddings with the outputs of the Hint Tuner ($\tilde{\mathbf{H}}_{s:e}$ and $\tilde{\mathbf{F}}$). Here, $s$ and $e$ denote the start and end indices of the selected window within the full series $\mathbf{x}_{1:T}$. The model then autoregressively predicts the next token in the explanation sequence. The loss function is the negative log-likelihood of the target sequence:
\[
\mathcal{L}_{\text{LM}} = - \sum_{i=1}^{N} \log P\big(y_i \mid y_{<i}, q, \tilde{\mathbf{H}}_{s:e}, \tilde{\mathbf{F}}\big).
\]
By optimizing this objective, the Hint Tuner learns to map temporal features from the frozen encoder into meaningful "soft prompts" that effectively guide the frozen LLM to generate accurate and relevant explanations for the given time-series window. During this phase, gradients update only the Hint Tuner and its associated parameters ($\mathbf{M}$ and $\mathbf{P}_{\text{fix}}$); both $f_{\theta}$ and all LLM parameters remain frozen.

\subsection{\methodname\ Hyperparameter Configuration}
\label{appx:hyperparameters}

This section provides a comprehensive specification of the hyperparameter configuration used in our \methodname\ framework. The architecture consists of two primary components: the time-series encoder and the hint tuner, each with distinct parameter settings optimized through systematic ablation studies.

\paragraph{Time-Series Encoder Configuration.} The time-series encoder $f_{\theta}$ employs a Transformer-based architecture with patch-based tokenization. The patch size $P$ is set to 16, enabling efficient processing of long sequences while preserving temporal granularity. The encoder utilizes $L=6$ transformer layers, each with a model dimension $d_{\text{model}}=512$ and $H=8$ attention heads, resulting in a head dimension $d_{\text{head}}=64$. The projection dimension $d_{\text{proj}}$ is configured to 256.

\paragraph{Hint Tuner Architecture.} The Perceiver-based hint tuner serves as the cross-modal alignment module between time-series representations and the frozen LLM embedding space. The prototype bank consists of $P=1024$ prototype embeddings derived from the LLM vocabulary through a learned linear mapping. The number of fixed task-prior tokens $K$ is set to 8, providing sufficient capacity for global task-level priors while maintaining computational efficiency. The cross-attention mechanism within the hint tuner employs 8 attention heads, matching the encoder configuration for architectural consistency.

\paragraph{Language Model Integration.} Our framework supports multiple LLM families with varying parameter scales. The experiments primarily utilize Qwen2.5-7B-Instruct as the base model, with the hidden dimension $d_h=4096$ for the 7B variant. The vocabulary size varies by model family, typically ranging from 32,000 to 152,000 tokens. Special tokens \texttt{<|local\_hint|>} and \texttt{<|fixed\_hint|>} are added to the vocabulary for hint injection, with corresponding token IDs dynamically assigned during initialization.

\paragraph{Training Configuration.} The two-phase training procedure employs distinct hyperparameter settings for each phase. During Phase I (encoder pretraining), the reconstruction loss weight $\lambda_{\text{recon}}$ and classification loss weight $\lambda_{\text{class}}$ are both set to 1.0, providing balanced supervision across objectives. The masking ratio for reconstruction is configured to 0.25, following established practices in self-supervised time-series learning. Phase II (hint tuner training) utilizes standard causal language modeling with a learning rate of $2 \times 10^{-4}$ and weight decay of 0.01, applied exclusively to the trainable parameters of the hint tuner and prototype bank.

\paragraph{Inference Parameters.} During generation, the model employs beam search with 5 beams and a repetition penalty of 1.15 to ensure diverse and coherent explanations. The maximum generation length is capped at 1000 tokens to accommodate detailed explanations while preventing excessive verbosity. No-repeat n-gram size is set to 3 to avoid repetitive patterns, and length penalty is maintained at 1.0 to balance explanation completeness with conciseness.

\section{Prompt Templates and Q\&A Exemplars}

\subsection{Prompt Templates}
\label{appx:prompt-templates}
\subsubsection{Question Generation Prompt}
\label{appx:question-generation-prompt}
\begin{lstlisting}[style=prompt]
System: You generate precise and relevant questions for time series anomaly detection.

User:
Generate a {question_type} question focused on anomaly detection for the window [ {window_start}, {window_end} ].

Context:
- Task: time series anomaly detection on a windowed segment
- The window {may / does} contain anomalies: {has_anomaly}
- Canonical tag (if available): {canonical_tag}
- Global information: {global_information}

Requirements:
1) Output ONLY the question text (no answers, no explanations).
2) Focus on anomaly identification and pattern analysis within the window.
3) Consider boundary effects near window edges.
4) Multiple choice (if applicable):
   - Provide exactly 4 options (A, B, C, D).
   - Options must be mutually exclusive, same style.
   - Include both normal and anomalous descriptions; avoid exact numeric values.
5) True/False (if applicable):
   - Make a specific statement about a potential anomaly pattern.
6) Open-ended (if applicable):
   - Ask about pattern evidence and reasoning for/against anomalies.
\end{lstlisting}

\subsubsection{Answer Generation Prompt}
\label{answer-generation-prompt}
\begin{lstlisting}[style=prompt]
System: You analyze time series patterns and generate concise answers.

User:
You are given a time series window [ {window_start}, {window_end} ], which belongs to a longer series.
Global information: {global_information}
Canonical tag (if available): {canonical_tag}
Anomaly description (if any): {anomaly_description}
Data [current_value(normal_value)]: [{data_str}]
Question: {question}

Constraints:
- Focus on the pattern of current_values; avoid relying on normal_values.
- Keep the answer concise (<= 150 words), pattern-first (e.g., sustained level change, volatility burst).
- MCQ: start with the correct option letter, then explanation (e.g.,  B) ...).
- True/False: start with True or False, then explanation.
- Do not quote exact numeric values; reason from shape, persistence, variability.
- If no anomaly, state it clearly with supporting evidence.
\end{lstlisting}

\subsection{Q\&A Exemplars}
\label{appx:qa-exemplars}
\subsubsection{Multiple Choice}
\label{appx:qa-mcq}
\begin{lstlisting}[style=qa]
Type: multiple_choice
Window: [668, 707]
Has anomaly: false
Anomaly descriptions: []

Question:
Which of the following best describes the anomaly detection results for the time series window from step 668 to 707?

A) The window contains a sudden spike suggesting an isolated anomaly near its center.
B) The window shows a consistent pattern without any signs of anomalous behavior.
C) The window includes a gradual drift at the boundary, indicating a possible transition to an anomaly.
D) The window exhibits irregular fluctuations that are characteristic of multiple overlapping anomalies.

Answer:
B) The window shows a consistent pattern without any signs of anomalous behavior.
The time series window from step 668 to 707 displays a steady, gradual decrease in values, with each step typically dropping by 0.01 or remaining constant for a brief interval before continuing the gradual trend. There are no sudden spikes, irregular fluctuations, or abrupt changes that would indicate an anomaly. The pattern is smooth and consistent throughout the window, suggesting stable behavior and the absence of any anomalous events within this interval.
\end{lstlisting}

\subsubsection{True/False}
\label{appx:qa-tf}
\begin{lstlisting}[style=qa]
Type: true_false
Window: [418, 458]
Has anomaly: true
Anomaly descriptions:
[A local continuous upward spike anomaly, featuring 5 consecutive spikes with amplitudes from 3.03 to 5.93]

Question:
True or False: The presence of a cluster of consecutive upward spikes within the window from step 418 to 458 suggests an anomalous deviation from the expected time series pattern.

Answer:
True. Within the window from step 418 to 458, there is a cluster of consecutive upward spikes-specifically, a sequence of five elevated values (2.26, 2.05, 2.05, 2.05, 0.71) that stand out sharply against the otherwise stable and low baseline pattern (around 0.7). This sudden, localized series of spikes deviates significantly from the expected behavior of the time series, which is generally flat and consistent. The presence of these consecutive upward anomalies suggests a notable and abrupt deviation from the normal pattern, indicating an anomalous event within this segment of the data.
\end{lstlisting}

\subsubsection{Open-ended}
\label{appx:qa-oes}
\begin{lstlisting}[style=qa]
Type: open_ended
Window: [294, 311]
Has anomaly: false
Anomaly descriptions: []

Question:
How would you assess the presence or absence of anomalies in the time series window from step 294 to 311, and what evidence from the data supports your conclusion regarding the normality or abnormality of this segment?

Answer:
There is no evidence of anomalies in the time series window from step 294 to 311. The values show a gradual decrease from 0.88 to 0.62, followed by a mild recovery back up to 0.73. This pattern reflects a smooth and consistent trend without sudden spikes, abrupt drops, or irregular fluctuations. The transitions between values are steady and do not indicate any unusual or unexpected behavior. Overall, the segment demonstrates normal and stable behavior within the observed window.
\end{lstlisting}

\section{Phase I TSAD results}

\paragraph{Datasets.} We benchmark nine public univariate time-series anomaly detection corpora: IOPS, MGAB, NAB, NEK, Power, SED, TODS, UCR, and YAHOO. These datasets span industrial KPIs, synthetic chaotic dynamics, web traffic, power consumption, network/system logs, and curated mixed-domain series. Each series is standardized via z-score normalization using training-set statistics. 

\paragraph{Baselines.} The comparison covers classical, deep generative, and foundation-style models. IForest~\citep{liu2008isolation} and LOF~\citep{breunig2000lof} serve as nonparametric unsupervised baselines. OmniAnomaly~\citep{su2019robust}, USAD~\citep{audibert2020usad}, and TranAD~\citep{tuli2022tranad} represent deep unsupervised anomaly detectors. DADA~\citep{shentu2024towards} and TSPulse~\citep{ekambaram2025tspulse}, MOMENT\_ZS~\citep{goswami2024moment}, Chronos~\citep{ansari2024chronos}, TimesFM~\citep{das2024decoder}, and Time\_MOE~\citep{xiaoming2025time} are pretrained time-series models evaluated in zero-shot or lightly adapted settings. \methodname\ denotes a semantic representation-based detector. Implementations follow public releases or authors' configurations when feasible.

\paragraph{Metrics.} We report three standard metrics. (i) PA-F1: the point-adjusted F1 that credits a hit if any point within an anomalous window is flagged, computed from precision and recall after point adjustment; threshold for PA-F1 is selected on a validation split when available, or via an unsupervised percentile heuristic on training scores. (ii) AUC-ROC: area under the receiver operating characteristic, threshold-free and aggregated over all test points. (iii) AUC-PR: area under the precision–recall curve, which is more informative under severe class imbalance. Higher values indicate better performance for all metrics.

\begin{table*}[h]
\centering
\footnotesize
\caption{Model Performance Comparison Across All Datasets}
\label{tab:univariate_results}
\resizebox{\textwidth}{!}{%
\begin{tabular}{l|l||cccccccccccc}
\hline
& & \multicolumn{7}{c}{Foundation Models} & \multicolumn{3}{c}{Deep Learning} & \multicolumn{2}{c}{Classical}  \\
\cline{3-14}
Dataset & Metric & \methodname\ & Chronos & DADA & MOMENT\_ZS & TS\_Pulse & Time\_MOE & TimesFM & OmniAnomaly & TranAD & USAD & IForest & LOF \\
\hline
\multirow{3}{*}{IOPS} & PA-F1 & 57.19 & \underline{64.17} & 55.52 & 44.76 & 50.59 & 41.58 & \textbf{87.89} & 53.50 & 53.04 & 41.24 & 15.31 & 41.38 \\
 & AUC-ROC & 71.13 & 72.16 & 78.61 & \textbf{82.67} & 49.71 & 61.87 & 70.12 & 80.89 & 69.44 & \underline{81.31} & 48.29 & 66.80 \\
 & AUC-PR & 18.13 & 25.65 & \underline{29.59} & 22.45 & 1.65 & 24.15 & 28.33 & \textbf{41.63} & 25.51 & 21.96 & 6.49 & 25.63 \\
\hline
\multirow{3}{*}{MGAB} & PA-F1 & 7.13 & 10.16 & 3.11 & 2.90 & \underline{11.39} & 2.64 & 7.74 & 7.62 & 7.80 & 3.49 & 5.85 & \textbf{12.31} \\
 & AUC-ROC & \textbf{69.97} & 50.78 & 54.12 & 47.47 & 49.94 & 37.73 & 49.97 & 60.15 & \underline{61.04} & 51.12 & 43.88 & 50.27 \\
 & AUC-PR & \underline{0.55} & 0.31 & 0.31 & 0.25 & 0.27 & 0.20 & 0.27 & 0.41 & 0.42 & \textbf{1.74} & 0.24 & 0.34 \\
\hline
\multirow{3}{*}{NAB} & PA-F1 & 93.73 & 97.08 & 96.18 & 94.61 & 97.76 & 90.48 & \textbf{99.09} & \underline{98.02} & 97.93 & 93.42 & 42.81 & 86.79 \\
 & AUC-ROC & 51.79 & 53.82 & 56.23 & \underline{64.01} & 50.28 & 50.82 & 53.33 & 56.26 & 54.33 & \textbf{73.09} & 55.26 & 49.93 \\
 & AUC-PR & 21.00 & 20.62 & 21.87 & \underline{42.73} & 13.91 & 20.32 & 21.12 & 24.87 & 21.92 & \textbf{53.14} & 20.48 & 18.55 \\
\hline
\multirow{3}{*}{NEK} & PA-F1 & 87.19 & 98.44 & \underline{98.68} & 81.10 & 88.14 & 50.82 & \textbf{98.79} & 88.50 & 86.78 & 66.26 & 72.47 & 87.35 \\
 & AUC-ROC & 53.28 & 57.60 & 82.51 & \textbf{92.02} & 53.94 & 37.41 & 62.64 & 88.85 & 81.25 & \underline{91.65} & 85.66 & 81.38 \\
 & AUC-PR & 30.32 & 30.63 & 44.90 & 55.75 & 16.34 & 7.13 & 34.05 & \textbf{73.10} & \underline{57.39} & 52.91 & 52.76 & 55.63 \\
\hline
\multirow{3}{*}{Power} & PA-F1 & \textbf{98.94} & 95.97 & 89.53 & 90.25 & \underline{98.77} & 83.29 & 97.58 & 86.66 & 87.51 & 68.06 & 0.00 & 87.92 \\
 & AUC-ROC & \underline{62.47} & 48.82 & 46.19 & 43.23 & 50.68 & 38.92 & 46.78 & 59.88 & 57.05 & \textbf{66.46} & 50.00 & 38.93 \\
 & AUC-PR & \textbf{20.52} & 10.44 & 10.07 & 10.68 & 11.09 & 9.01 & 9.98 & 13.64 & 12.37 & \underline{17.30} & 10.97 & 8.98 \\
\hline
\multirow{3}{*}{SED} & PA-F1 & \textbf{93.59} & 38.77 & 18.69 & 1.92 & \underline{70.26} & 33.29 & 32.49 & 2.25 & 4.49 & 0.00 & 31.62 & 0.00 \\
 & AUC-ROC & \textbf{96.35} & 53.03 & 14.85 & 1.43 & 49.23 & \underline{69.08} & 29.64 & 28.93 & 12.89 & 10.71 & 27.96 & 27.35 \\
 & AUC-PR & \textbf{73.76} & 4.80 & 2.77 & 2.57 & 4.95 & \underline{7.45} & 3.26 & 3.35 & 2.78 & 2.68 & 3.26 & 3.25 \\
\hline
\multirow{3}{*}{TODS} & PA-F1 & 77.50 & \textbf{83.40} & 61.20 & 26.86 & 53.03 & 44.92 & \underline{81.26} & 38.82 & 40.31 & 40.86 & 33.25 & 54.62 \\
 & AUC-ROC & \textbf{92.40} & 68.45 & 68.16 & 46.17 & 50.20 & 51.68 & \underline{72.96} & 49.69 & 48.87 & 53.92 & 51.16 & 53.12 \\
 & AUC-PR & \textbf{68.20} & 35.60 & 20.33 & 10.03 & 6.98 & 10.36 & \underline{36.69} & 6.89 & 6.74 & 17.37 & 7.43 & 19.74 \\
\hline
\multirow{3}{*}{UCR} & PA-F1 & 47.58 & 43.71 & 34.90 & 25.55 & \underline{56.57} & 31.63 & \textbf{60.88} & 31.44 & 32.67 & 22.57 & 23.02 & 32.43 \\
 & AUC-ROC & \textbf{79.34} & 55.79 & 56.27 & 54.92 & 50.57 & 55.38 & \underline{57.21} & 50.67 & 50.64 & 53.74 & 51.93 & 55.86 \\
 & AUC-PR & \textbf{24.06} & 6.11 & 2.19 & 6.92 & 0.85 & 2.12 & 6.35 & 2.91 & 2.32 & \underline{8.08} & 2.54 & 2.50 \\
\hline
\multirow{3}{*}{YAHOO} & PA-F1 & 53.88 & \underline{54.09} & 50.82 & 12.54 & 7.99 & 21.39 & \textbf{87.18} & 18.66 & 8.09 & 7.40 & 2.86 & 29.78 \\
 & AUC-ROC & \textbf{96.55} & 94.64 & 95.95 & 74.37 & 49.02 & 63.08 & \underline{96.35} & 77.71 & 56.47 & 62.54 & 48.90 & 72.85 \\
 & AUC-PR & \textbf{86.07} & 70.75 & 76.66 & 4.84 & 1.49 & 22.08 & \underline{79.23} & 16.28 & 2.90 & 3.72 & 1.66 & 44.26 \\
\hline
\multirow{1}{*}{Overall} & Avg. Rank & \textbf{3.81} & 4.74 & 5.52 & 7.48 & 7.70 & 8.70 & \underline{4.41} & 5.48 & 7.00 & 6.78 & 9.07 & 7.30 \\
\hline
\end{tabular}
}%
\end{table*}

\paragraph{Results.} The experimental results, summarized in Table~\ref{tab:univariate_results}, demonstrate that our proposed \methodname\ model achieves competitive performance across the nine public benchmarks. When compared against a variety of SOTA methods, including recent time-series foundation models, \methodname\ consistently ranks as a top performer on several datasets such as SED, TODS, UCR, and YAHOO, achieving the best results on multiple metrics. This validates the effectiveness of our approach for time-series anomaly detection and shows its potential to match or even exceed the current state of the art.

\section{Detailed Procedure for G-eval}
\label{appx:g-eval}
This appendix documents the evaluation methodology, criteria, and prompts used by our G-Eval variant.

\subsection{Methodological Overview}

We adopt an advanced G-Eval judge that evaluates system answers along multiple dimensions and calibrates discrete scores using token-level log-probabilities. For each question-answer instance:
\begin{enumerate}
  \item Determine the question type (e.g., multiple choice, open-ended, true/false) and instantiate the corresponding set of evaluation dimensions.
  \item For each dimension, query a LLM judge with a criterion-specific instruction (see the exact prompt in Appx.) to produce a reasoning-based assessment and a final integer score \(s \in \{1,2,3,4,5\}\). If score extraction is ambiguous, a conservative default is used.
  \item Obtain token-level probabilities assigned to the score tokens \(\{1,\dots,5\}\). Exponentiate and normalize these probabilities to construct a score distribution. Use its expectation as a log-probability-weighted score and compute a confidence via normalized entropy (Appx.~\ref{sec:gev-adv-weighting}).
\end{enumerate}

\subsection{Evaluation Criteria and Weights}

For each \texttt{question\_type}, the evaluator applies a fixed set of dimensions with 5-point guidelines and combines them via type-specific weights:
\begin{table}[h]
\centering
\caption{Dimension Weights by Question Type.}
\label{tab:gev-dim-weights}
\begin{tabular}{l|l}
\hline
\textbf{multiple\_choice} & correctness: 0.70, reasoning\_quality: 0.30 \\
\textbf{open\_ended}      & relevance: 0.30, completeness: 0.35, accuracy: 0.35 \\
\textbf{true\_false}      & correctness: 0.60, justification\_quality: 0.40 \\
\hline
\end{tabular}
\end{table}

The dimension descriptions and scoring guidelines (1--5) are:

\paragraph{multiple\_choice}
\begin{itemize}
  \item \textbf{correctness}: How accurate is the generated response compared to the expected answer?
    \begin{itemize}
      \item 5: Perfect match, completely correct
      \item 4: Mostly correct; minor deviations not affecting core meaning
      \item 3: Partially correct; captures some key aspects but misses important details
      \item 2: Somewhat relevant but with significant errors or omissions
      \item 1: Incorrect or completely irrelevant
    \end{itemize}
  \item \textbf{reasoning\_quality}: How well does the response demonstrate logical reasoning and explanation?
    \begin{itemize}
      \item 5: Clear, logical, comprehensive reasoning fully explains the choice
      \item 4: Good reasoning with minor gaps
      \item 3: Adequate reasoning; lacks depth or has inconsistencies
      \item 2: Weak reasoning; significant gaps or flawed logic
      \item 1: No clear reasoning or completely flawed logic
    \end{itemize}
\end{itemize}

\paragraph{open\_ended}
\begin{itemize}
  \item \textbf{relevance}: How relevant and on-topic is the generated response?
    \begin{itemize}
      \item 5: Completely relevant; directly addresses all aspects
      \item 4: Highly relevant; minor omissions
      \item 3: Moderately relevant; addresses core aspects but misses details
      \item 2: Somewhat relevant; off-topic content or key omissions
      \item 1: Irrelevant or off-topic
    \end{itemize}
  \item \textbf{completeness}: How complete and comprehensive is the response?
    \begin{itemize}
      \item 5: Fully comprehensive; covers all necessary aspects
      \item 4: Mostly complete; minor gaps
      \item 3: Adequately complete; missing some important details
      \item 2: Incomplete; significant information missing
      \item 1: Very incomplete
    \end{itemize}
  \item \textbf{accuracy}: How factually accurate is the response?
    \begin{itemize}
      \item 5: Completely accurate; no factual errors
      \item 4: Mostly accurate; very minor inaccuracies
      \item 3: Generally accurate; some notable errors
      \item 2: Several factual errors affecting reliability
      \item 1: Major factual errors or mostly inaccurate
    \end{itemize}
\end{itemize}

\paragraph{true\_false}
\begin{itemize}
  \item \textbf{correctness}: How correct is the T/F judgment and supporting explanation?
    \begin{itemize}
      \item 5: Perfect judgment; excellent supporting explanation
      \item 4: Correct judgment; good explanation
      \item 3: Correct judgment; adequate explanation
      \item 2: Incorrect judgment; reasonable attempt at explanation
      \item 1: Incorrect judgment; poor or no explanation
    \end{itemize}
  \item \textbf{justification\_quality}: How well does the response justify the decision?
    \begin{itemize}
      \item 5: Excellent justification with clear evidence and reasoning
      \item 4: Good justification with solid evidence
      \item 3: Adequate justification with some evidence
      \item 2: Weak justification with little evidence
      \item 1: No meaningful justification
    \end{itemize}
\end{itemize}

\subsection{Advanced Scoring From Log-Probabilities}
\label{sec:gev-adv-weighting}

Let \(s\in\{1,2,3,4,5\}\) denote the raw score extracted from the LLM's evaluation output. We also obtain token-level log-probabilities for score tokens \(\{1,\dots,5\}\). The evaluator constructs a distribution over scores by exponentiating and normalizing these log-probabilities:
\[
p(s) \propto \exp(\log p_s),\quad 
\tilde{p}(s) = \frac{\exp(\log p_s)}{\sum_{k=1}^{5} \exp(\log p_k)}.
\]
The \textbf{weighted score} is the expectation under \(\tilde{p}\):
\[
\mathrm{WeightedScore} = \sum_{s=1}^{5} s \cdot \tilde{p}(s).
\]

\subsection{Advanced G-Eval Prompt}

\begin{lstlisting}[style=prompt]
You are an expert evaluator for natural language generation systems. Your task is to evaluate the quality of generated responses using the G-Eval methodology with chain-of-thought reasoning.

Control the Maximum Length to 500 words.

**Evaluation Criterion: {criterion.dimension}**
{criterion.description}

**Scoring Guidelines:**
{guidelines_text}

**Question:** {question}

**Expected Answer:** {expected_answer}

**Generated Response:** {generated_response}

**Instructions:**
1. Analyze the generated response step by step using chain-of-thought reasoning
2. Compare it against the expected answer for the specified criterion
3. Consider both content quality and alignment with the expected answer
4. Provide detailed reasoning for your evaluation
5. Conclude with a single score from 1-5
6. Ignore error in index mismatch, just focus on the content

Please follow this exact format for your response:

**Step-by-step Analysis:**
[Provide detailed chain-of-thought analysis]

**Comparison with Expected Answer:**
[Compare generated response with expected answer]

**Final Assessment:**
[Summarize your evaluation]

**Score:** [Single integer: 1, 2, 3, 4, or 5]
\end{lstlisting}

\subsection{Detailed results of G-eval}
\begin{figure}[htb]
    \centering
    \includegraphics[width=\linewidth]{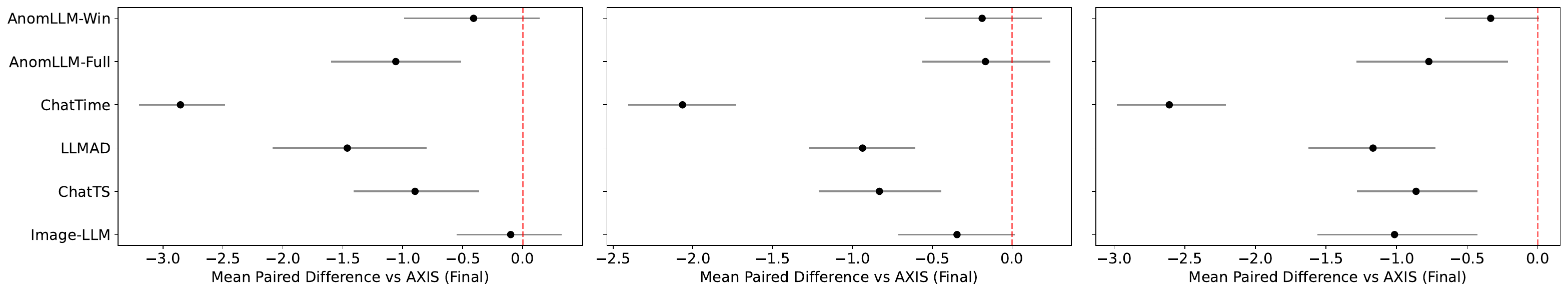}
    \caption{Comparative analysis of baseline models against AXIS via forest plot. Each horizontal line represents the 95\% bootstrap confidence interval for the mean paired score difference relative to the AXIS baseline. The central dot marks the point estimate of the mean difference. The vertical red dashed line at zero indicates no difference; confidence intervals crossing this line suggest that the model's performance is not statistically different from the baseline.}
    \label{fig:forest-table1}
\end{figure}

\begin{figure}[htb]
    \centering
    \includegraphics[width=\linewidth]{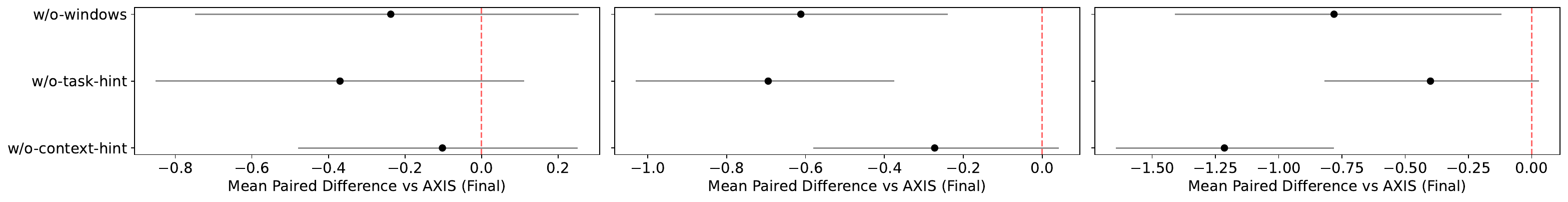}
    \caption{Forest plot comparing \methodname\ variants against the AXIS baseline. The plot displays the mean paired score difference and the 95\% bootstrap confidence interval for each variant. The vertical red line at zero indicates no performance difference relative to AXIS.}
    \label{fig:forest-table2}
\end{figure}

\begin{figure}[htb]
    \centering
    \includegraphics[width=\linewidth]{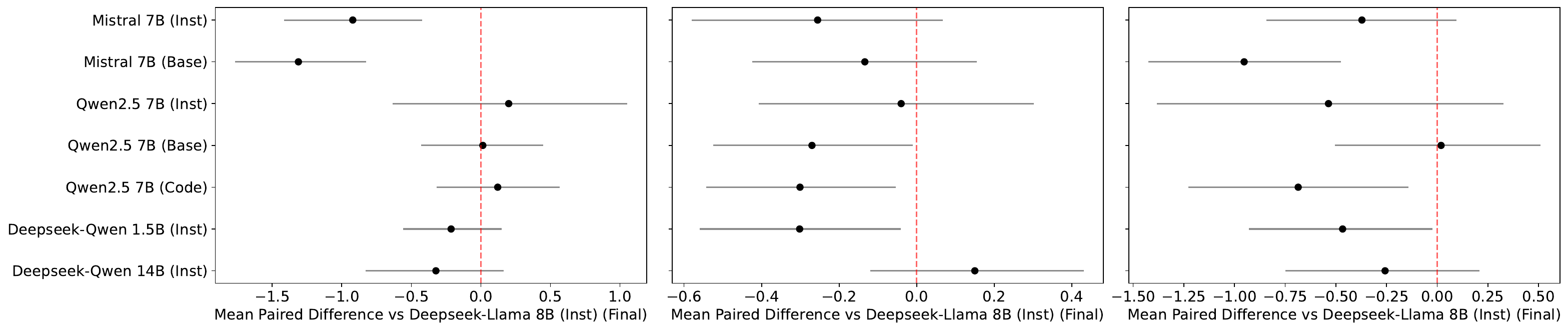}
    \caption{Forest plot for ablation studies, benchmarked against AXIS. This plot illustrates the mean paired score difference and the corresponding 95\% bootstrap confidence interval for each ablated variant. The vertical red line at zero serves as the reference for no difference from the baseline.}
    \label{fig:forest-table3}
\end{figure}

Figures~\ref{fig:forest-table1} to \ref{fig:forest-table3} present forest plots to visually summarize the pairwise model performance comparisons from Tables~\ref{tab:main-table} to \ref{tab:variants}. The confidence intervals are calculated from the variance of the pairwise total scores.

As shown in Fig.~\ref{fig:forest-table1}, our AXIS model demonstrates a strong performance. On Multiple-choice questions, it significantly outperforms all models except for Image-LLM and Anon-LLM(Window). For True/False questions, AXIS significantly surpasses all other models, highlighting its advantage in objective answering. Furthermore, on Open-Ended questions, AXIS outperforms other models while achieving performance comparable to the General LLM (AnomLLM).

Fig.~\ref{fig:forest-table2} details an ablation study of our proposed AXIS model. The results demonstrate that AXIS consistently and significantly outperforms its variants ("w/o-windows", "w/o-task-hint", and "w/o-context-hint") across all three question categories. The confidence intervals for the mean paired difference are entirely below zero in all subplots, indicating that the removal of any of these components leads to a statistically significant degradation in performance. This underscores the integral contribution of each component to the overall efficacy of the AXIS model.

Fig.~\ref{fig:forest-table3} provides a comparative analysis of various open-source models against the Deepseek Llama 8B (Inst) baseline. In multiple-choice questions, Deepseek Llama 8B shows a significant performance advantage over the Mistral 7B models, while its performance is statistically comparable to the Qwen2.5 7B series. For open-ended and true/false questions, Deepseek Llama 8B shows comparable performance to most Qwen and Deepseek-Qwen variants, showing that the robustness of our AXIS design.

\section{Procedure and detail results for Human-based evaluation}
\label{appx:human}
\subsection{Designs of questionnaire}

Our human expert questionnaire is systematically designed to evaluate model explanations across multiple dimensions. Each questionnaire includes: (1) \textbf{Question Information}: the original question, expected answer, and question type classification (multiple choice, open-ended, or true/false); (2) \textbf{Time Series Visualization}: a plot showing the full time series with the target window highlighted to provide visual context; (3) \textbf{Model Responses}: all baseline model outputs presented in a randomized order to minimize bias; (4) \textbf{Evaluation Criteria}: dimension-specific scoring guidelines adapted from G-Eval methodology, with 5-point scales for correctness, reasoning quality, relevance, completeness, accuracy, and justification quality depending on question type; (5) \textbf{Scoring Tables}: structured evaluation forms for each criterion; (6) \textbf{Model Ranking}: comparative ranking of all models with justifications. 

For our study, a total of 140 unique questions were assessed. To ensure robust and unbiased evaluation, we collected 280 completed questionnaires, ensuring that each question was independently evaluated by two human experts under a double-blind protocol. This comprehensive design guarantees systematic and unbiased human evaluation, capturing both quantitative scores and rich qualitative insights from domain experts.

\begin{tcolorbox}[colback=gray!5, colframe=gray!90, title=Human Expert Evaluation Questionnaire Example]

\textbf{Question 1}

\textbf{Question Type:} multiple\_choice

\textbf{Question:}
Which of the following best describes the likelihood of detecting an anomaly within the time series window from step 491 to 521, given its observed behavior?

A) There is a sudden spike at the start of the window that indicates potential anomalous activity  
B) The window displays periodic irregularities that suggest subtle anomalies  
C) The window exhibits consistent and stable behavior, suggesting no detectable anomalies  
D) The pattern shows gradual drift, implying an evolving anomaly towards the window's end

\textbf{Expected Answer:}
C) The window exhibits consistent and stable behavior, suggesting no detectable anomalies. The current\_value data fluctuates around zero with both positive and negative values but does not display any abrupt spikes, sustained trends, or irregular periodicities...

\textbf{Time Series Visualization:} [Plot showing full time series with highlighted window]

\hrule

\textbf{Model Responses:}

\textbf{Model 1}
To determine anomaly likelihood in the window from step 491 to 521, let's examine the values: [0.12, -2.45, -1.15, ...]. Within this window, there is a sharp drop from 0.12 to -2.45 at the very start... **A) There is a sudden spike at the start of the window that indicates potential anomalous activity**

\textbf{Model 2}
Within the specified window (indices 491 to 521), the data shows fluctuating values without any abrupt spikes or consistent irregular patterns...

\textbf{Model 3}
The observed behavior within the time series window from step 491 to 521 indicates no sudden spikes or abrupt changes... **C) The window exhibits consistent and stable behavior, suggesting no f anomalies.**

\textbf{Model 4}
The time series visualization highlights the analysis window from step 491 to 521... **C) The window exhibits consistent and stable behavior, suggesting no detectable anomalies.**

\textbf{Model 5}
\{``briefExplanation'': \{``step1\_global'': ``The overall trend of the latest 31 data points is stable...'', ``is\_anomaly'': false, ``anomalies'': [], ...\}

\textbf{Model 6}
**Answer: C) The window exhibits consistent and stable behavior, suggesting no detectable anomalies.** The time series window from step 491 to 521 shows a mix of positive and negative values with moderate fluctuations, but there are no sudden spikes...

\hrule

\textbf{Evaluation Criteria:}

\textbf{Correctness (1-5):} How accurate is the generated response compared to the expected answer?
\textbf{Reasoning Quality (1-5):} How well does the response demonstrate logical reasoning and explanation?

\textbf{Scoring Table:}
\begin{tabular}{|l|c|c|c|}
\hline
Model & Correctness & Reasoning Quality & Comments \\
\hline
Model 1 & \_\_\_ & \_\_\_ & \_\_\_ \\
Model 2 & \_\_\_ & \_\_\_ & \_\_\_ \\
Model 3 & \_\_\_ & \_\_\_ & \_\_\_ \\
Model 4 & \_\_\_ & \_\_\_ & \_\_\_ \\
Model 5 & \_\_\_ & \_\_\_ & \_\_\_ \\
Model 6 & \_\_\_ & \_\_\_ & \_\_\_ \\
\hline
\end{tabular}

\textbf{Model Ranking:} Rank all models from best (1) to worst (6) with justifications.

\end{tcolorbox}

\subsection{Detailed results for Human evaluation}
\begin{figure}[h]
    \centering \includegraphics[width=0.8\linewidth]{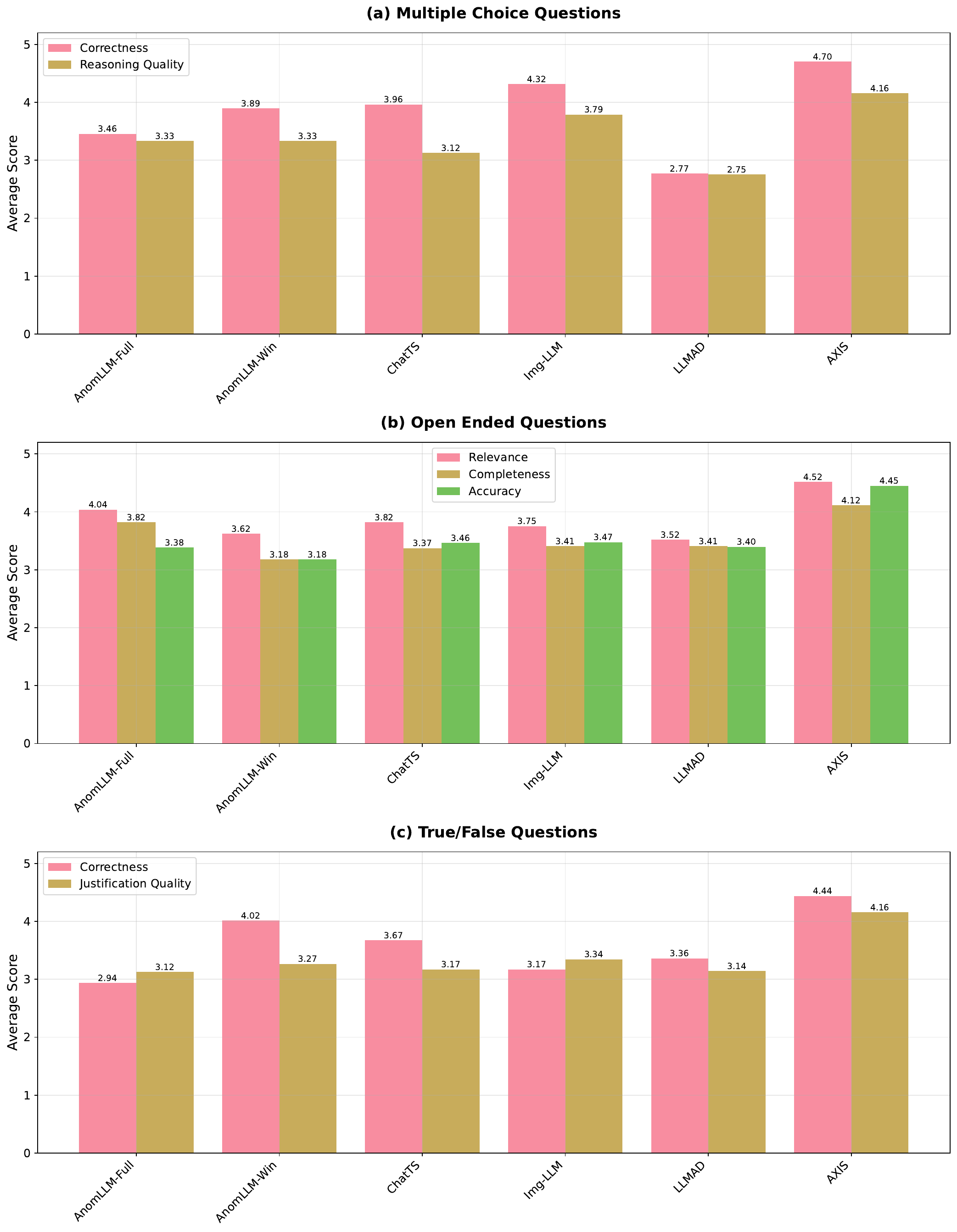}
    \caption{The mean scores by human evaluation for (a) multiple choice question, (b) open ended questions and (c) true/false questions}
    \label{fig:detail-human}
\end{figure}

Fig.~\ref{fig:detail-human} provides a criterion-level breakdown of human evaluation across the three question types. \methodname\ ranks first consistently. On Multiple Choice questions, it achieves the best scores in both \emph{Correctness} and \emph{Reasoning Quality}, with a clear margin over the next best visual baseline. This indicates that our hint-based conditioning not only selects the right option but also articulates concise, logically grounded rationale.

For Open-Ended questions, \methodname\ leads on \emph{Relevance}, \emph{Completeness}, and \emph{Accuracy}, reflecting faithful, fully supported explanations rather than surface descriptors. On True/False questions, it also tops both \emph{Correctness} and \emph{Justification Quality}, showing strong calibration and evidence-backed decisions. Overall, these results demonstrate that the proposed three-pathway design (symbolic numeric grounding, context-integrated local hints, and task-prior hints) confers robust advantages across formats—surpassing specialized TS-LLMs and multimodal vision–language approaches in both accuracy and human-judged explanatory quality.

\section{Causal Ablation Analysis for Hint Tokens}
\label{appx:causal-ablation}

To analyze the individual contribution of each local hint token to the model's generation quality, we implement a causal ablation study. This method systematically removes or modifies specific hint tokens to quantify their importance in producing accurate explanations.

Given hint embeddings $\tilde{\mathbf{H}}_{s:e} \in \mathbb{R}^{(e-s) \times d_h}$, we measure the contribution of each local hint token at position $i$ within the target window $[s,e)$ through the following procedure:

\paragraph{Baseline Computation.} First, we compute the baseline log-likelihood using the complete hint embeddings:
\[
\mathcal{L}_{\text{baseline}} = -\log P(\mathbf{y} | \tilde{\mathbf{H}}_{s:e}, q, \tilde{\mathbf{F}}),
\]
where $\mathbf{y}$ is the target explanation, $q$ is the query, and $\tilde{\mathbf{F}}$ represents the task-prior hints.

\paragraph{Ablation Methods.} For each position $i \in [s,e)$, we create an ablated version of the hint embeddings $\tilde{\mathbf{H}}_{s:e}^{(i)}$ using \textbf{Zero Replacement:} $\tilde{\mathbf{H}}_{s:e}^{(i)}[i-s] = \mathbf{0}$.

\paragraph{Contribution Score.} The contribution score for position $i$ is computed as:
\[
C_i = \mathcal{L}_{\text{baseline}} - \mathcal{L}_{\text{ablated}}^{(i)},
\]
where $\mathcal{L}_{\text{ablated}}^{(i)} = -\log P(\mathbf{y} | \tilde{\mathbf{H}}_{s:e}^{(i)}, q, \tilde{\mathbf{F}})$. A positive $C_i$ indicates that removing the hint at position $i$ degrades the model's performance, suggesting that this position contributes positively to explanation generation.

\begin{figure}
    \centering
    \includegraphics[width=0.8\linewidth]{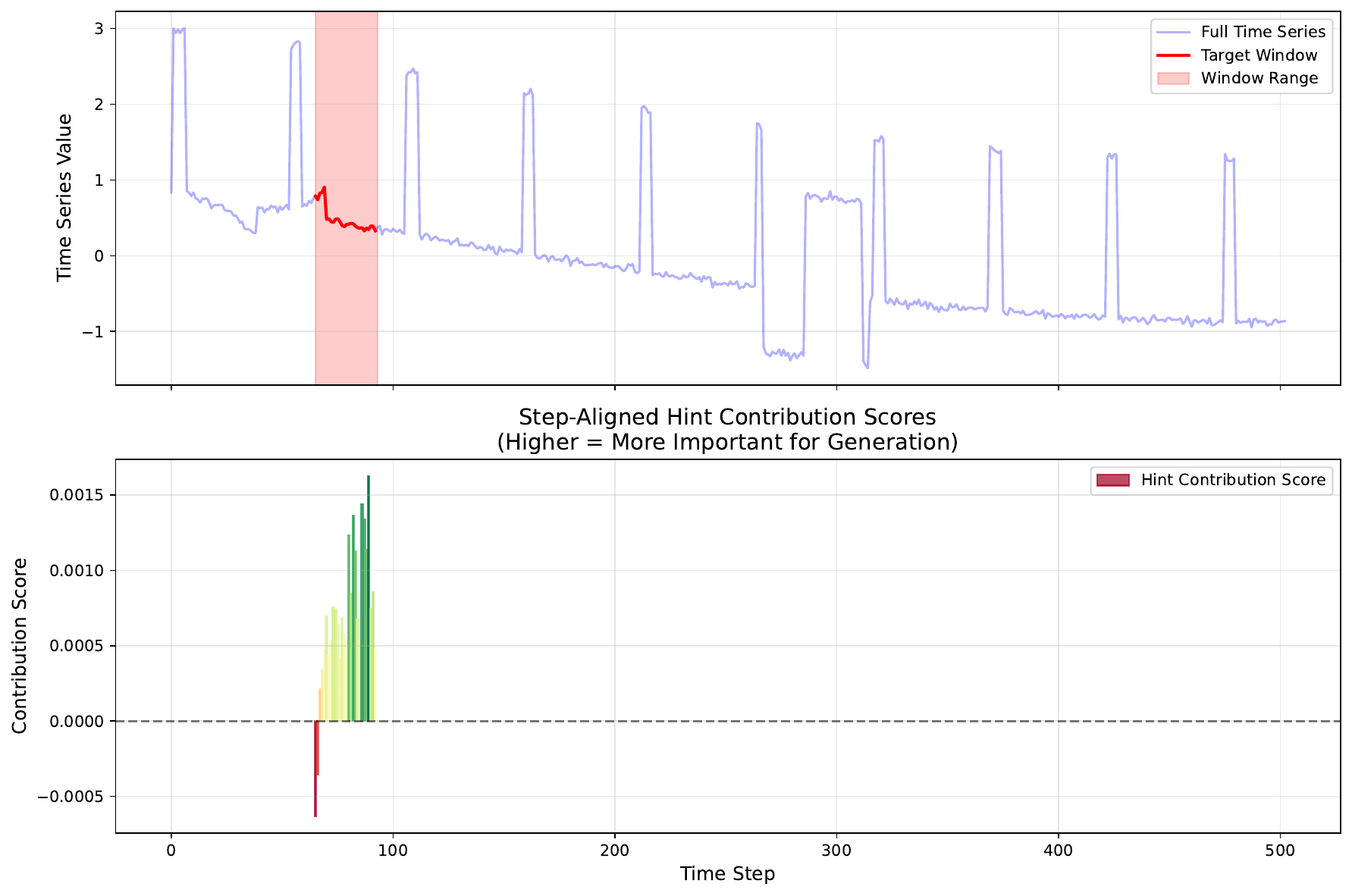}
    \caption{Causal Ablation Analysis of Step-Aligned Hint Tokens. The figure visualizes the contribution score $C_i$ for each hint token, computed via our causal ablation method, alongside the original input time series.}
    \label{fig:step-aligned-1}
\end{figure}

\begin{figure}
    \centering
    \includegraphics[width=0.8\linewidth]{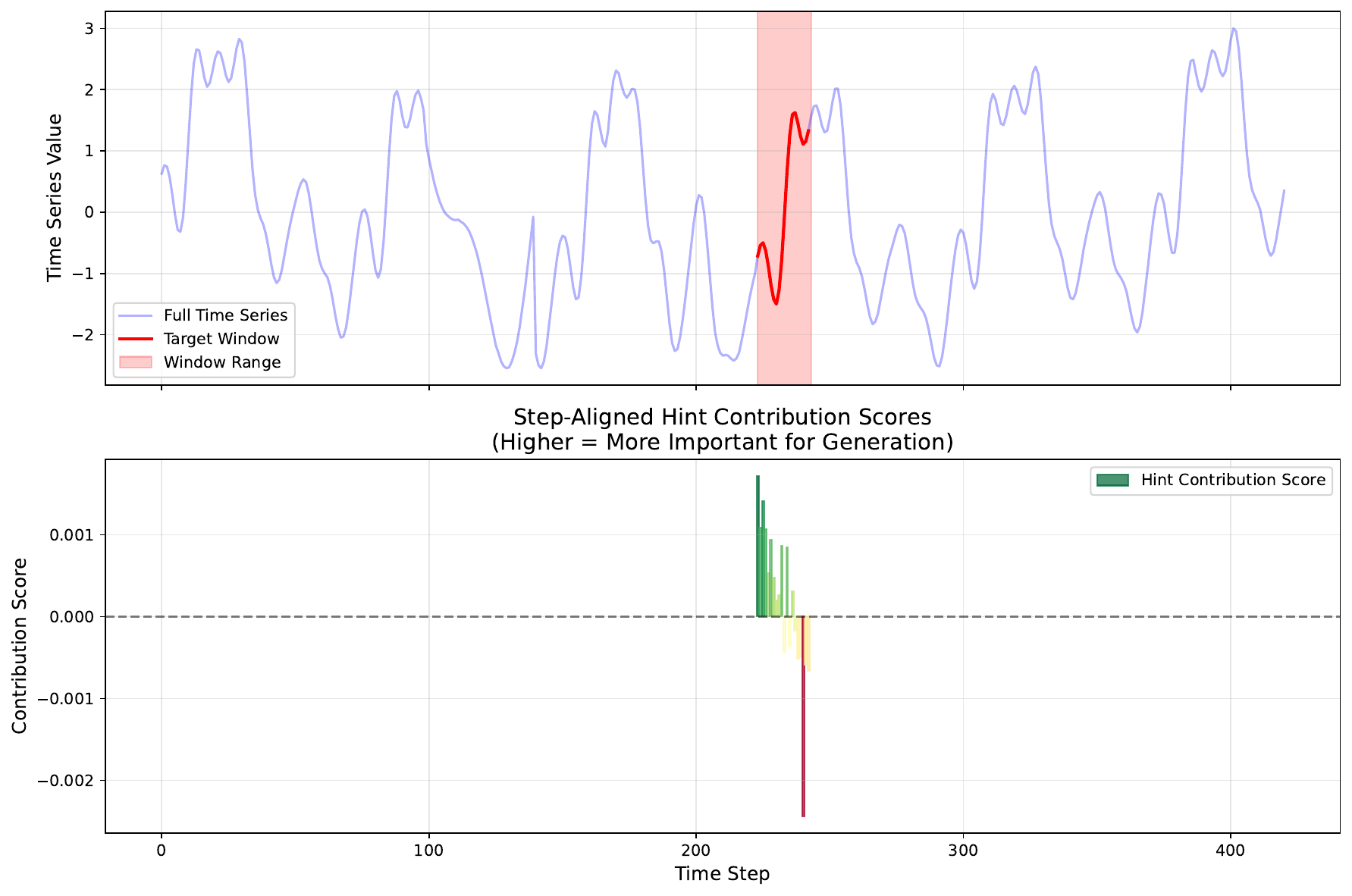}
    \caption{Causal Ablation Analysis of Step-Aligned Hint Tokens. The figure visualizes the contribution score $C_i$ for each hint token, computed via our causal ablation method, alongside the original input time series.}
    \label{fig:step-aligned-2}
\end{figure}

\paragraph{Validation of Context-Integrated Hints via Causal Ablation Analysis.}
As depicted in Fig.~\ref{fig:step-aligned-1} and Fig.~\ref{fig:step-aligned-2}, the causal ablation analysis provides empirical evidence for the efficacy of our proposed hint mechanism. A predominant observation across both examples is that the majority of the contribution scores, denoted by $C_i$, are positive ($C_i > 0$). According to the formulation presented in Appx.~\ref{appx:causal-ablation}, a positive $C_i$ indicates that ablating the hint token at position $i$ degrades the model's performance, as measured by an increase in the negative log-likelihood of the target explanation. This finding strongly supports our hypothesis that the context-integrated, step-aligned hints furnish valuable guidance for the model, making a net positive contribution to the generation of high-quality answers.

Furthermore, a more nuanced pattern emerges from the results. We observe that in regions where the time series exhibits significant volatility or sharp fluctuations, the corresponding contribution scores are comparatively lower. This suggests that the inherent complexity and unpredictability of volatile segments in the time series can diminish the marginal utility of individual hint tokens. In essence, while the hints remain beneficial overall, their directional impact is partially mitigated by the increased difficulty of the task in these challenging temporal regions.


\end{document}